\makeatletter
\newcommand*{\addFileDependency}[1]{
  \typeout{(#1)}
  \@addtofilelist{#1}
  \IfFileExists{#1}{}{\typeout{No file #1.}}
}
\makeatother

\newcommand*{\myexternaldocument}[1]{%
    \externaldocument{#1}%
    \addFileDependency{#1.tex}%
    \addFileDependency{#1.aux}%
}

\documentclass[10pt,twocolumn,letterpaper]{article}

\usepackage[pagenumbers]{cvpr} 

\usepackage{graphicx}
\usepackage{amsmath}
\usepackage{amssymb}
\usepackage{booktabs}
\usepackage{xr}
\usepackage{caption}
\usepackage{subcaption}
\usepackage{float}

\usepackage{hyperref}

\usepackage[capitalize]{cleveref}
\crefname{section}{Sec.}{Secs.}
\Crefname{section}{Section}{Sections}
\Crefname{table}{Table}{Tables}
\crefname{table}{Tab.}{Tabs.}

\def\cvprPaperID{9880} 
\def\confName{CVPR}
\def\confYear{2022}

\usepackage{overpic}
\usepackage{enumitem} 
\usepackage{overpic} 
\usepackage{color}

\definecolor{turquoise}{cmyk}{0.65,0,0.1,0.3}
\definecolor{purple}{rgb}{0.65,0,0.65}
\definecolor{dark_green}{rgb}{0, 0.5, 0}
\definecolor{orange}{rgb}{0.8, 0.6, 0.2}
\definecolor{red}{rgb}{0.8, 0.2, 0.2}
\definecolor{darkred}{rgb}{0.6, 0.1, 0.05}
\definecolor{blueish}{rgb}{0.0, 0.3, .6}
\definecolor{light_gray}{rgb}{0.7, 0.7, .7}
\definecolor{pink}{rgb}{1, 0, 1}
\definecolor{greyblue}{rgb}{0.25, 0.25, 1}

\usepackage{blindtext}

\renewcommand{\paragraph}[1]{\vspace{1em}\noindent\textbf{#1}.}

\usepackage{multirow}
\usepackage{xr}
\usepackage{caption}
\usepackage{subcaption}
\begin{document}
\title{Topology-Preserving Shape Reconstruction and Registration via Neural Diffeomorphic Flow}

\author{Shanlin Sun, Kun Han, Deying Kong, Hao Tang, Xiangyi Yan, Xiaohui Xie\\
University of California, Irvine\\
{\tt\small \{shanlins, khan7, deyingk, htang6, xiangyy4, xhx\}@uci.edu}\\
}
\maketitle

\begin{abstract}
Deep Implicit Functions (DIFs) represent 3D geometry with continuous signed distance functions learned through deep neural nets. Recently DIFs-based methods have been proposed to handle shape reconstruction and dense point correspondences simultaneously, capturing semantic relationships across shapes of the same class by learning a DIFs-modeled shape template.
These methods provide great flexibility and accuracy in reconstructing 3D shapes and inferring correspondences.
However, the point correspondences built from these methods do not intrinsically preserve the topology of the shapes, unlike mesh-based template matching methods.  This limits their applications on 3D geometries where  underlying topological structures exist and matter, such as anatomical structures in medical images. 
In this paper, we propose a new model called Neural Diffeomorphic Flow (NDF) to learn deep implicit shape templates, representing shapes as conditional diffeomorphic deformations of templates, intrinsically preserving shape topologies.
The diffeomorphic deformation is realized by an auto-decoder consisting of Neural Ordinary Differential Equation (NODE) blocks that progressively map shapes to implicit templates.
We conduct extensive experiments on several medical image organ segmentation datasets to evaluate the effectiveness of NDF on reconstructing and aligning shapes. NDF achieves consistently state-of-the-art organ shape reconstruction and registration results in both accuracy and quality. The source code is publicly available at \url{https://github.com/Siwensun/Neural_Diffeomorphic_Flow--NDF}.
\end{abstract}

\begin{figure}[!ht]
\begin{center}
\includegraphics[width=\columnwidth]{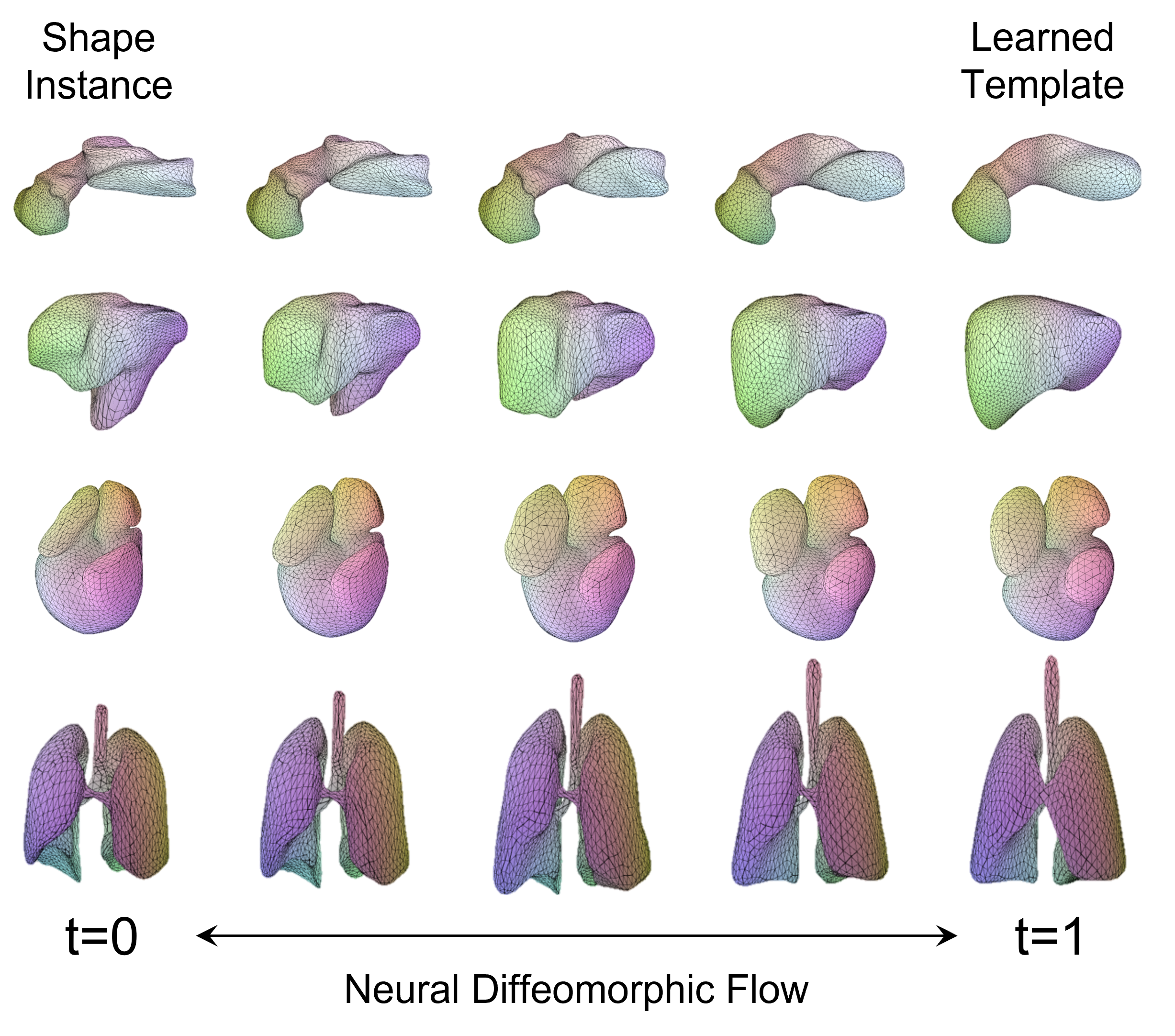}
\end{center}
\caption{Our model deforms back and forth between shape instances and learned templates through neural diffeomorphic flow, in a invertible and progressive manner. Colors on the shape surface show the correspondences and edges show the topologies.
}
\label{fig:teaser}
\end{figure}
\section{Introduction}
\label{sec:intro}

3D geometry representation is fundamental to many downstream tasks in computer vision such as 3D model understanding, reconstruction, and matching. In particular, shape representation is of vital importance to many medical image applications such as organ segmentation \cite{tang2021recurrent,yan2022after,chen2021deep,tang2021spatial, you2020unsupervised, you2021momentum, you2021simcvd}, medical image reconstruction \cite{li2019novel, guha2020deep, you2019ct}, shape abnormality detection and surgical navigation \cite{suinesiaputra2017statistical, lehmann2009integrating}. 

Recently deep implicit functions (DIFs) have emerged as an effective and efficient tool for modeling 3D objects \cite{mescheder2019occupancy, park2019deepsdf, chen2019learning, sitzmann2020implicit, xu2019disn}. Compared to traditional representations such as voxel grids, point clouds, and polygon meshes, DIF-based 3D representations have the advantages of being compact while at the same time enjoying strong representation power, making it more suitable for modeling complex shapes with fine geometric details. However, DIFs do come with a strong drawback - it is difficult to establish correspondences between two shapes, unlike the traditional method such as meshes. This drawback limits the application of DIFs for shape analysis, especially in many medical image applications, where being able to map and compare shapes is often a necessity. 

A number of methods have been proposed to address the limitation of DIFs. The DIT (Deep Implicit Templates) \cite{zheng2021deep} and DIF-Net (Deformed Implicit Field), build upon DeepSDF \cite{park2019deepsdf}, formulates DIFs as conditional deformations of a template deep implicit function, and uses a spatial warping module to explicitly model the conditional deformations and infer point-wise transformations. \cite{liu2020learning} learns dense 3D shape correspondence from semantic part embedding by introducing an inverse implicit function to BAE-Net \cite{chen2019bae}. 

However, a common drawback of the above methods is that the conditional deformation modeled by these methods (e.g., LSTM in DIT) is agnostic of the topology of the shapes. This will be problematic in situations where two shapes share the same topology and we want the topology to be preserved after deformation. Applications falling into this category include modelling 3D shapes of human body, anatomical structures in medical images, and other objects with fixed topologies. What's more, considering the small number of anatomical shapes available for training, it is challenging to generalize the learned deformations to unseen data if no shape prior is utilized.

In this work, we propose a new formulation of DIFs called Neural Diffeomorphic Flow (NDF) for representing 3D shapes. Similar to DIT and DIF-Net, NDF models shapes as conditional deformations of a template DIF. But different from DIT, the conditional deformation is intrinsically diffeomorphic, thereby ensuring that the resulting deformation is topology preserving. NDF is also different from AtlasNet \cite{groueix2018papier}, which can preserve topology but requires predefined fixed topology as its shapes are modeled by meshes.

Our main contributions are summarized as follows: 
\begin{itemize}
    \item We introduce invertible NDF to match a shape to its implicit template. It can align point clouds or meshes without sacrificing accuracy while guaranteeing topology preservation. 
    \item We design a quasi time-varying velocity field to learn diffeomorphic flows based on neural ODEs, allowing us to model shape deformation in a progressive and time-invertible manner. 
    %
    
    \item 
    We tested NDF on multiple organ datasets and demonstrated that it leads to state-of-the-art shape reconstruction and registration results on both existing and new shapes. On shape registration, NDF generates one or several order of magnitude fewer unpleasant faces . 
    
\end{itemize}
\begin{figure}
\centering
    \begin{subfigure}[b]{\columnwidth}
        \centering
        \includegraphics[height=0.5\columnwidth]{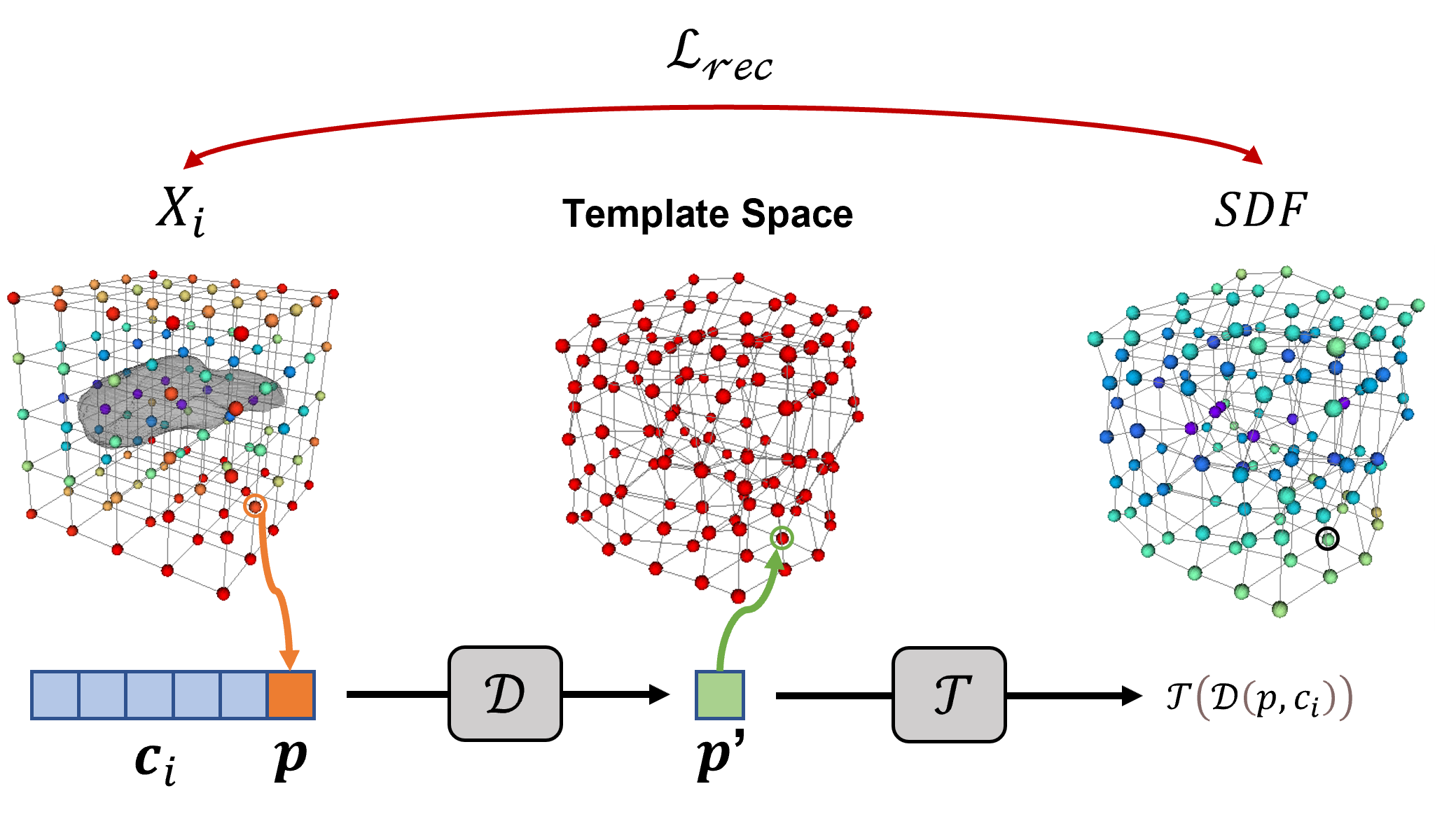}
        \caption{Train}
        \label{fig:overview_train}
    \end{subfigure}
    \vfill
    \begin{subfigure}[b]{\columnwidth}
        \centering
        \includegraphics[height=0.6\columnwidth]{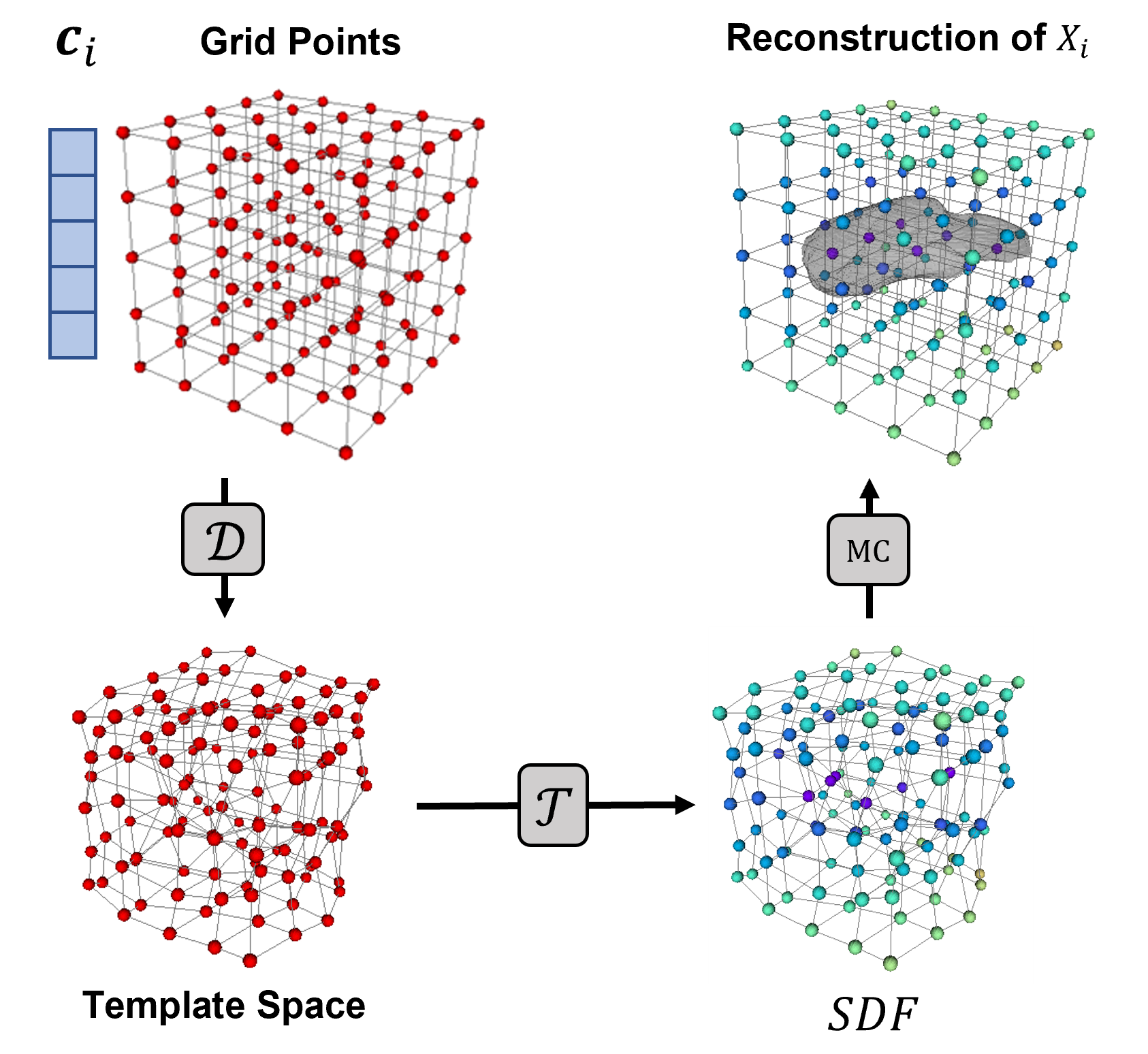}
        \caption{Reconstruction}
        \label{fig:overview_recon}
    \end{subfigure}
    \vfill
    \begin{subfigure}[b]{\columnwidth}
        \centering
        \includegraphics[height=0.3\columnwidth]{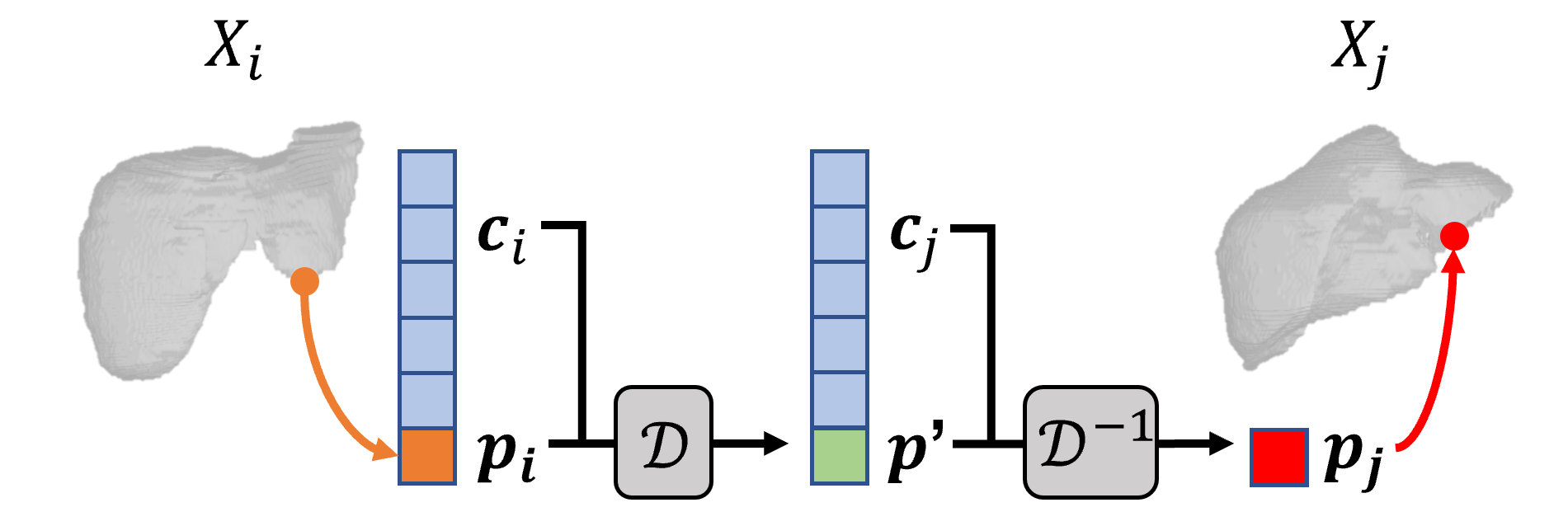}
        \caption{Point Correspondence}
        \label{fig:overview_pc}
    \end{subfigure}
    \caption{\textbf{Overview of NDF} - (a): We train deform code $\boldsymbol{c}_i$, deformation module $\boldsymbol{D}$ and deep sdf reprsentations $\boldsymbol{T}$ jointly. $\boldsymbol{p}$ is a sampled 3D position and $\boldsymbol{p}'$ is the deformation position of $\boldsymbol{p}$ in the template space. \textit{We sample points from the continuous 3D space, but we draw grids here only for illustration}; (b): Suppose we have already optimized $\boldsymbol{c}_i$, to reconstruct an unseen shape, we input grid points into our model and the reconstructed mesh is obtained via marching cube (MC) model output; (c): illustration of point correspondences as in Eq.\ref{eq.point_correspondence}.}
    \label{fig:overview}
\end{figure}

\section{Related works}
\label{sec:related}

\paragraph{Deep Implicit Functions}
Traditional implicit function is defined in the grid space and extracts the explicit shape surface from its zero-level set. Deep implicit function is the extension of traditional implicit function to represent shapes in continuous 3D space and have shown great representation capacity. DeepSDF \cite{park2019deepsdf} is an example of auto-decoder models representing continuous SDF. Many works are developed based on it, among which \cite{jiang2020local, chabra2020deep, tretschk2020patchnets} try to depict finer structures by modelling SDF in the unit of local regions. Additionally, DualSDF \cite{hao2020dualsdf} designs dual pathways (primitive and accurate) to represent SDF with VAD framework and C-DeepSDF \cite{duan2020curriculum} intends to improve the training strategy via curriculum learning. Occupancy Network \cite{mescheder2019occupancy} represents another branch of deep implicit function that constructs the solid mesh via classifying 3D points whether are included in mesh or not. Occupancy Flow \cite{niemeyer2019occupancy} follows similar ideas of Occpancy Network in shape representation but extend it to 4D with a continuous vector field in time and space.

\paragraph{Point Correspondence and Shape Registration}
There are several ways to achieve point correspondence such as template learning \cite{groueix2018papier, huang2015analysis, kim2013learning, loper2015smpl, wang2018pixel2mesh}, elementary representation \cite{genova2019learning, genova2019deep, deprelle2019learning}, deformation field-based methods \cite{luthi2017gaussian, myronenko2010point} and so on. Mesh-based templates are very popular in representing similar shapes such human body, face and hand, where the templates topology are fixed, so they cannot deal with topological changes. Element-based methods can only capture structure-level features because they aim to describe complex shapes with simple elements. DIF-Net \cite{deng2021deformed} and DIT \cite{zheng2021deep} are typical deformation field-based methods and DIT can generate smoother deformation because it applies LSTM to do deformations. Our work can be seen as deformation field-based methods but our deformation field is topology-preserving and invertible. 

\paragraph{Diffeomorphic Transformation}
A diffeomorphism is an invertible mapping where the forward and backward transformations are smooth. It is widely used in nonrigid registration and shape analysis. Model complexity and computation makes it hard to be incorporated with deep learning solution. \cite{detlefsen2018deep} is the first paper that built diffeomorphic image transformations into a deep classification model via CPAB transformations \cite{freifeld2017transformations}. In terms of medical registration problem, people usually assume the velocity field is stationary and defined in the grid space \cite{ashburner2007fast}. Many works \cite{dalca2018unsupervised, balakrishnan2019voxelmorph, krebs2019learning, dalca2019unsupervised, avants2008symmetric} used scaling and squaring method \cite{arsigny2006log} to do fast integration of stationary velocity field. 
Recently, with the power of neural ordinary differential equation (NODE) solver \cite{chen2018neuralode, chen2021eventfn}, optimizing a neural diffeomorphic flow efficiently became possible. Occupancy Flow \cite{niemeyer2019occupancy}, based on Occupany Network, learns 4D reconstruction with implicit correspondences by modelling a temporally and spatially continuous vector field. Neural Mesh Flow\cite{gupta2020neural} focuses on generating manifold mesh from images or point clouds via conditional continuous diffeomorphic flow. PointFlow \cite{yang2019pointflow} incorporates continuous normalizing flows with a principle probabilistic framework to reconstruct 3d point clouds.
\section{Method}

Our Deep Implicit Function representation via NDF follows the formulation of Deep Implicit Templates (DIT) \cite{zheng2021deep}, which decompose a Coded Shape DeepSDF \cite{park2019deepsdf} into a Single Shape DeepSDF and a conditional spatial warping function. Our work depicts this spatial warping function as a conditional diffeomorphic flow under which topology is preserved. In this section, we first review Deep Implicit Templates and then introduce our proposed conditional spatial deformation.

\subsection{Review of Deep Implicit Templates}
\label{sec.dit_review}
Deep Implicit Templates, same as DeepSDF, represent a 3D shape $X_i$ with a continuous signed distance field (SDF) $\mathcal{F}$. Given a random 3D point $p$ and deform code $\boldsymbol{c}_i$ of length $k$, $\mathcal{F}$ outputs the point's distance to the closest surfaces, whose sign indicates whether the point lies inside or outside the underlying shape surface:
\begin{equation}
    \mathcal{F}(\boldsymbol{p}, \boldsymbol{c}_i)=s, \quad\mathrm{where\ }\boldsymbol{p} \in \mathbb{R}^{3}, \boldsymbol{c}_i \in \mathcal{R}^k, s \in \mathbb{R}
    \label{eq.sdf_overview}
\end{equation}

During training, each shape code is paired with one training shape $X_i$.  During inference, the deform code corresponding to a new shape is obtained via optimization. The underlying shape surface is implicitly expressed as the zero-level set surface of $\mathcal{F}$, obtained with, for example, Marching Cubes \cite{lorensen1987marching}.

Different from DeepSDF where the physical meaning of latent code $\boldsymbol{c}$ is ambiguous, DIT treats $\boldsymbol{c}$ as a variable controlling how each shape deforms to a template shape, so that the conditional continuous SDF $\mathcal{F}$ can be decompose into $\mathcal{T} \circ \mathcal{D}$
\begin{equation}
    \mathcal{F}(\boldsymbol{p}, \boldsymbol{c}_i)=\mathcal{T}(\mathcal{D}(\boldsymbol{p}, \boldsymbol{c}_i))
    \label{eq.dit_overview}
\end{equation}

where $\mathcal{D}: \mathbb{R}^{3} \times \mathbb{R}^{k} \mapsto \mathbb{R}^{3}$ is the conditional spatial deformation module that maps the coordinate of $p$ of shape $X_i$ to a canonical position $p'$ given deform code $\boldsymbol{c}_i$ and $\mathcal{T}$ is essentially a single shape DeepSDF modeling the implicit template. By this design, it builds up point correspondences between the learned template and each shape instance, on top of which correspondences across all shapes within one category are achieved.

\subsection{Neural Diffeomorphic Flow}
\label{sec:ddit}
\paragraph{Diffeomorphic Flow}
We intend to establish dense point correspondences between each shape object and the template shape, and keep the desired geometric topology using diffeomorphic flow. Let $\Phi_i(\boldsymbol{p}, t): \mathbb{R}^{3} \times [0, 1] \rightarrow \mathbb{R}^{3}$ describe the continuous trajectory of a 3D point ($\boldsymbol{p}$) during the time interval [0, 1] where the starting points and destination points respectively located in the SDF of shape $X_i$ and the template shape. And let $\mathbf{v}_i(\boldsymbol{p}, t): \mathbb{R}^{3} \times [0, 1] \rightarrow \mathbb{R}^{3}$ define the velocity field of 3D points with respect to shape $X_i$ in time interval $[0, 1]$. The diffeomorphic flow $\Phi_i$ of shape $X_i$ is the solution of the initial value problem (IVP) of an ordinary differential equation (ODE) as below:
\begin{equation}
    \frac{\partial \Phi_i}{\partial t}(\boldsymbol{p}, t)=\mathbf{v}_i(\Phi_i(\boldsymbol{p}, t), t) \quad \text { s.t. } \quad \Phi_i(\boldsymbol{p}, 0)=\boldsymbol{p}
    \label{eq.forward_ode}
\end{equation}

where $\mathbf{p} \in \mathbb{R}^{3}$ is a 3D position on the SDF of shape $X_i$. Thus, the diffeomorphic deformations module $\mathcal{D}$ conditioning on the deform code of shape $X_i$ is given by: 
\begin{gather}
    \mathcal{D}(\boldsymbol{p}, \boldsymbol{c}_i) = \Phi_i(\boldsymbol{p}, 1) \label{eq.deformation_module}
\end{gather}

If  velocity field $\mathbf{v}_i(\cdot, \cdot)$ is globally Lipschitz continuous, the solution for the IVP exists and is unique in the interval $[0, 1]$, which means any two ODE trajectories do not cross each other \cite{dupont2019augmented}. This can provide the diffeomorphic flow with the property of topology preservation to maintain structure consistency.  

Diffeomorphic flow is invertible, thus the inverse flow $\Psi_i: \mathbb{R}^{3} \times[0, 1] \rightarrow \mathbb{R}^{3}$ from template shape to shape $X_i$ can be obtained by solving the following ODE:
\begin{gather}
    \frac{\partial \Psi_i}{\partial t}(\boldsymbol{p}, t)=-\mathbf{v}_i(\Psi_i(\boldsymbol{p}, t), t) \quad \text { s.t. } \quad \Psi_i(\boldsymbol{p}, 0)=\boldsymbol{p}
    \label{eq.backward_ode} \\
    \mathcal{D}^{-1}(\boldsymbol{p}, \boldsymbol{c}_i) = \Psi_i(\boldsymbol{p}, 1) \label{eq.inverse_deformation_module}
\end{gather}

where $\mathbf{p}$ is a 3D point on the SDF of template shape and $\mathcal{D}^{-1}$ denotes the inverse diffeomorphic field. So far, the invertible deformation between any shape instance and the templates of its category can be described as the integral of the velocity field. 

\paragraph{Conditional Quasi Time-varying Velocity Field}
Our goal is to learn a neural network that parameterizes the velocity field to capture the dense topology-preserved point correspondence across shape objects. In this section, we will describe how we design it. We denote $\mathbf{v}$ as the neural network representing the velocity field in time and space, and $\mathbf{v}_i$ describes the neural velocity field with respect to the shape $X_i$, i.e., $\mathbf{v}_i(\boldsymbol{p}, t) = \mathbf{v}(\boldsymbol{p}, \boldsymbol{c}_i, t)$, where $\boldsymbol{c}_i$ a vector controls how neural velocity field deforms points in the SDF of $X_i$. 

$\mathbf{v}_i(\boldsymbol{p}, t)$ is a general expression of velocity field depending on both time and position, here we call it time-varying velocity field to distinguish with stationary velocity field $\mathbf{v}_i(\boldsymbol{p})$, where the velocity of a point in the field only decided by its position. In our case, training time-varying velocity field might be difficult because unlike 4D reconstruction \cite{niemeyer2019occupancy} having adequate training samples of multiple frames, our model can only be supervised on $t=0$ and $t=1$. In dealing with the medical registration problem defined in regular grid space, people often assume the deformable transformation is based on a stationary velocity field because it can be efficiently integrated through scaling and squaring technique \cite{higham2005scaling, moler2003nineteen}. 

Our insight is to design a quasi time-varying velocity field composed of several subsequent stationary velocity fields to realize progressive deformations. Concretely, we hope the first several diffeomorphic flows are able to roughly align the shape instance to its template shape while the last several diffeomorphic flows take the minor adjustment in geometric details. Suppose this quasi time-varying velocity field is made up with $K$ stationary velocity fields. Let $\mathbf{v}_{i}^{k}(\cdot): \mathbb{R}^{3} \rightarrow \mathbb{R}^{3}$ describe the k-th stationary velocity field of shape $X_i$ and $\chi_A(\cdot)$ is an indicator function of $A$. The velocity field is governed by the following formula
\begin{equation}
    \mathbf{v}_i(\Phi_i(\boldsymbol{p}, t), t)=\sum_{k=0}^{K} \chi_{\left[ \frac{k}{K},\frac{k+1}{K}\right]}(t) \cdot \mathbf{v}_{i}^{k}(\Phi_i(\boldsymbol{p}, t))
    \label{eq.progressive_v_field}
\end{equation}

As shown in Eq.\ref{eq.progressive_v_field}, the quasi time-varying velocity field is basically a step function regarding time $t$. We can further derive the diffeomorphic flow by integrating $\mathbf{v}_i(\cdot, \cdot)$
\begin{equation}
    \Phi_i(\boldsymbol{p}, t) = \Phi_i(\boldsymbol{p}, \frac{k}{K}) + \int_{\frac{k}{K}}^{t-\frac{k}{K}} \mathbf{v}_{i}^{k}(\Phi_i(\boldsymbol{p}, t)) \mathrm{d}t
    \label{eq.progressive_d_field}
\end{equation}

where $t \in \left[\frac{k}{K}, \frac{k+1}{K} \right]$ and $k \in \{0, 1, ..., K-1\}$. This equation can be solved with a neural ordinary differential equation (NODE) solver \cite{chen2018neuralode}. In other words, $\Phi_i(\boldsymbol{p}, t)$ is the output of a NODE block receiving $\mathbf{v}(\boldsymbol{p}, \boldsymbol{c}_i, t)$ as dynamic function. To make this conditional velocity field neural network compatible with NODE, the conditional parameters (deform codes $\boldsymbol{c}$) stay unchanged when solving the integral.

We use residual MLP architecture similar to \cite{niemeyer2019occupancy, gupta2020neural} to represent the velocity field. For each shape $X_i$, we initialize the deform code $\boldsymbol{c}_i$ as \cite{park2019deepsdf} suggests, which can be either concatenated or multiplied to the point features. 

\subsection{Training}
We employ two modules to represent continuous SDF: a conditional deformation module $\mathcal{D}$ and a single shape DeepSDF $\mathcal{T}$. Like the other auto-decoder models, these two modules and deform code $\boldsymbol{c}$ are trained jointly (as illustrated in Fig.~\ref{fig:overview_train}) with a reconstruction loss and a regularization loss:
\begin{equation}
    \mathcal{L} = \mathcal{L}_{rec} + \lambda_{reg} \mathcal{L}_{reg}
    \label{eq.loss_overview}
\end{equation}

Since the deformation between a shape and the template shape performs in a progressive manner, we choose to use the curriculum learning strategy same as \cite{zheng2021deep, duan2020curriculum}. \cite{duan2020curriculum} set different curriculum learning hyper-parameters at different training stages and \cite{zheng2021deep} set different hyper-parameters for different warping stages. In our work, we will count the deformations of different timestamps into curriculum learning. To this end, the reconstruction loss can be written as:
\begin{equation}
    \mathcal{L}_{r e c}=\sum_{t \in T} \sum_{i=1}^{N} \sum_{j=1}^{S} L_{\epsilon_{t}, \lambda_{t}}\left(\mathcal{T}\left(\Phi_i(\boldsymbol{p}_j, t)\right), s_{i, j}\right)
    \label{eq.recon_loss}
\end{equation}

where $T$ is the set of evaluating timestamps, $N$ is the number of training shapes, S is the number of SDF samples for one shape, $s_{i, j}$ is the ground truth SDF of the j-th samples point $\boldsymbol{p}_j$ from the i-th shape and $\Phi_i(\boldsymbol{p}, t)$ is defined as in Eq.\ref{eq.progressive_d_field}. $L_{\epsilon_{t}, \lambda_{t}}$ is the curriculum training loss where $\epsilon$ controls the width of the tolerance zone and $\lambda$ controls the importance of the hard and semi-hard examples. The evaluation timestamps $T$ is set to be $\{0.25,  0.5, 0.75, 1.0\}$ in practice. For more details about curriculum learning, please refer to \cite{duan2020curriculum}.

Our regularization loss is very concise because NODE solver can largely prevent self-intersection \cite{dupont2019augmented,zhang2020approximation} with no explicit regularization. 
In other words, we don't need to design the point pair regularization \cite{zheng2021deep} or deformation smoothness prior \cite{deng2021deformed} to avoid the local distortion. In \cite{gupta2020neural}, the authors showed a toy example to demonstrate the "regularfzizar's delemma" that introducing strong regularization in training might lead to an unpleasant mesh reconstruction results. In our setting, we only need to constrain the magnitude of deformation field as well as learned deform codes. 
\begin{equation}
    \mathcal{L}_{reg}=\sum_{t \in T} \sum_{i=1}^{N} \sum_{j=1}^{S} \mathcal{L}_{0.25}\left(\left\|\Phi_i(\boldsymbol{p}_j, t)-\boldsymbol{p}_j\right\|_{2}\right) + \sum_{k=1}^{K}\left\|\boldsymbol{c}_{k}\right\|_{2}^{2}
    \label{eq.pw_loss}
\end{equation}
where $\mathcal{L}_{0.25}$ is the Huber loss with $\delta=0.25$. The goal of this point-wise regularization loss is to prevent the model from learning an over-simplified template but to look for a template shape which owns the most common structures that all shape instances within one category share.

\subsection{Inference}
At inference, after fixing the trained parameters of NDF, a deform code $\boldsymbol{c}_i$ for a new shape $X_i$ should be obtained via optimization in the first place. In our work, shape reconstruction is to extract the zero-level set surface from the SDF of an unseen shape object, which is predicted by our model given the optimized deform code. The correspondence between two shape objects $X_i$ and $X_j$ can be found via forward diffeomorphic deformations from $X_i$ to template space and then backward diffeomorphic deformations from template space to $X_j$, given $\boldsymbol{c}_i$ and $c_j$ respectively.

\paragraph{Learn deform code}
Same as DeepSDF \cite{park2019deepsdf}, a deform code $\boldsymbol{c}_i$ of shape $X_i$ is the Maximum-a-Posterior estimation as:
\begin{equation}
    \hat{\boldsymbol{c}_i}=\underset{\boldsymbol{c}_i}{\arg \min } \sum_{\left(\boldsymbol{p}, s\right) \in X_i} \ell\left(\mathcal{F}(\boldsymbol{p}, \boldsymbol{c}_i), s\right)+\frac{1}{\sigma^{2}}\|\boldsymbol{c}_i\|_{2}^{2}
    \label{eq.learn_latent_code}
\end{equation}

Different from the progressive reconstruction loss we designed for training, $\ell(\cdot, \cdot)$ here is the absolute error between model outputs and ground truth. 

\paragraph{Reconstruction}
Having learned the deform code $\boldsymbol{c}_i$, shape $X_i$ comes from the zero-level set surface of $\mathcal{F}(\boldsymbol{p}_{grid}, \boldsymbol{c}_i)$ (shown in Fig.~\ref{fig:overview_recon}), where $\boldsymbol{p}_{grid} \in \mathbb{Z}_{+}^3$. The resolution of reconstructed mesh can be manipulated by the number of grid points. In practice, we sample $256^3$ grid points for all deep implicit functions for comparison. 

\paragraph{Point correspondence and Shape Registration}
\label{sec:pc and sr}
In AtlasNet \cite{groueix2018papier}, the points that could find correspondence are only template mesh vertices. As for implicit template-based methods such as \cite{deng2021deformed, zheng2021deep}, they learn dense point correspondence, which is approximated since it is built by nearest neighbour search in the template space. 

In our work, suppose we have a 3D point $\boldsymbol{p}_i$ in shape $X_i$, its correspondence point $\boldsymbol{p}_j$ in shape $X_j$ can be found by (also shown in Fig.~\ref{fig:overview_pc}):
\begin{equation}
    \boldsymbol{p}_j = \mathcal{D}^{-1}\left(\mathcal{D}(\boldsymbol{p}_i, \boldsymbol{c}_i),\boldsymbol{c}_j\right)
    \label{eq.point_correspondence}
\end{equation}

where $\boldsymbol{c}_i$ and $\boldsymbol{c}_j$ are the optimized deform codes of shape $X_i$ and $X_j$ respectively.

Compared to point correspondence, shape registration not only seeks the aligned point set but also the aligned mesh. That means, given the source mesh $\mathcal{M}_s$ with vertices $\mathcal{V}_s$ and edges $\mathcal{E}_s$, the target aligned mesh $\mathcal{M}_t$ is $(\mathcal{V}_t, \mathcal{E}_s)$, where $\mathcal{V}_t$ is the correspondence point set of $\mathcal{V}_s$. 


\myexternaldocument{sec/3_method}

\section{Experiments}
\paragraph{Datasets}
NDF focuses on reconstructing shapes with common intrinsic topology, we choose to demonstrate our results on four medical datasets: Pancreas CT \cite{holger2016pancreas}, Multi-Modality Whole Heart Segmentation \cite{zhuang2016multi}, Lung and Liver, since these four types of organ have clear common topology but are of fair shape variation as can be learned from the difference between learned templates and shape instances. For more details about data source and preparation, please refer to supplementary material.


\paragraph{Experimental Setup}
We conduct two types of experiments to support the effectiveness of NDF. First, we investigate the representation power of our diffeomorphic flow-based methods on training samples and the reconstruction power on unseen shapes. We then evaluate the quality of the learned correspondence between two meshes (shape registration). 

The natural baselines for shape registration are DIT\cite{zheng2021deep} and DIF-Net \cite{deng2021deformed} because we share the similar shape representation formula based on deep implicit function. We also compare our model with AtlasNet \cite{groueix2018papier} which reconstructs shape using explicit mesh parameterization. 
%
%
To make our comparisons fair, we build our model based on the implementation of DeepSDF and choose it as the baseline for 3D shape representation basides DIT, DIF-Net and AtlasNet. 

\subsection{Shape Representation and Reconstruction}
\label{sec.shape reconstruction}

\begin{table*}
\centering
\resizebox{\linewidth}{!}{%
\begin{tabular}{lcccccccccccccccc} 
\toprule
               & \multicolumn{4}{c}{CD Mean ($\downarrow$)} & \multicolumn{4}{c}{CD Median ($\downarrow$)} & \multicolumn{4}{c}{NC Mean ($\uparrow$)} & \multicolumn{4}{c}{NC Median ($\uparrow$)}  \\ 
\cmidrule(lr){2-5}\cmidrule(lr){6-9}\cmidrule(lr){10-13}\cmidrule(lr){14-17}
Model / Organ          & Pancreas & Liver    & Lung     & Heart    & Pancreas & Liver    & Lung     & Heart    & Pancreas & Liver    & Lung     & Heart           & Pancreas & Liver    & Lung     & Heart             \\ 
\midrule
AtlasNet\_Sph \cite{groueix2018papier} & 8.08   & 3.46  & 5.01  & 7.55  & 7.44  & 2.46  & 3.76  & 7.38  & 0.703   & 0.823 & 0.824 & 0.808           & 0.7     & 0.829 & 0.826 & 0.814              \\
AtlasNet\_25 \cite{groueix2018papier}  & 6.05   & 2.48  & 207   & 4.86  & 5.64  & 1.72  & 4.54  & 3.86  & 0.65    & 0.818 & 0.772 & 0.824           & 0.643   & 0.823 & 0.791 & 0.823               \\
DeepSDF \cite{park2019deepsdf}      & 0.711  & 0.539 & 0.669 & \textbf{0.951} & 0.675 & 0.536 & 0.661 & \textbf{0.898} & 0.898   & 0.866 & 0.928 & 0.913           & 0.903   & 0.868 & 0.929 & 0.92              \\
DIF-Net \cite{deng2021deformed}          & 4.18   & 1.58  & 1.86  & 2.23  & 3.97  & 1.25  & 1.67  & 1.83  & 0.756   & 0.832 & 0.882 & 0.838           & 0.768   & 0.837 & 0.885 & 0.838              \\
DIT \cite{zheng2021deep}          & 0.63   & 0.509 & 0.712 & 1.05  & 0.658 & 0.505 & 0.693 & 0.976 & 0.903   & 0.87  & 0.934 & 0.919           & 0.904   & 0.873 & 0.934 & 0.93               \\
\midrule
\textbf{Ours}           & \textbf{0.512} & \textbf{0.476} & \textbf{0.643} & 0.993 & \textbf{0.515} & \textbf{0.479} & \textbf{0.631} & 0.925 & \textbf{0.917}   & \textbf{0.873} & \textbf{0.937} & \textbf{0.923}           & \textbf{0.918}   & \textbf{0.875} & \textbf{0.937}  & \textbf{0.932}              \\
\bottomrule
\end{tabular}
}
\caption{
\textbf{Shape Reconstruction} -- We demonstrate the reconstruction results of different representation methods on four organ categories. 
AtlasNet\_Sph and AtlasNet\_25 are AtlasNet using 3D sphere mesh and 25 square patches as the template shape respectively. The chamfer distance results shown above are multiplied by $10^3$. $\uparrow$ means higher is better and $\downarrow$ means lower is better. Here we use "Lung" to denote the union shape of lung and trachea organ and use "Heart" to denote the union structures of blood cavities.
} 
\label{tab:reconstruction}
\end{table*}
We use chamfer distance (CD) and normal consistency (NC) as the matrices to evaluate the quality of shapes representations and reconstructions by all methods. In the scope of this work, shape representation is to represent seen shape objects given the trained deform code and shape reconstruction is to represent unseen shape instances after optimizing the deform code. So, shape representation tells the effectiveness of representation methods while shape reconstruction reflects the generability. Due to space limitation, we only report the complete shape representation results for all datasets in supplementary material. 

Generally speaking, DIF-Net shows the strongest shape representation ability but very poor shape reconstruction performance. We believe the overfitting is a result of its point sampling strategy that many surface points are involved in training and inference. In Tab.~\ref{tab:reconstruction}, we can observe NDF achieves the best reconstruction results in terms of both chamfer distance and normal consistency for almost all datasets, compared to the state-of-the-art methods.

\subsection{Shape Registration}
\label{sec.shape registration}

\begin{figure}
\centering
    \begin{subfigure}[b]{0.49\columnwidth}
        \centering
        \includegraphics[height=0.6\columnwidth]{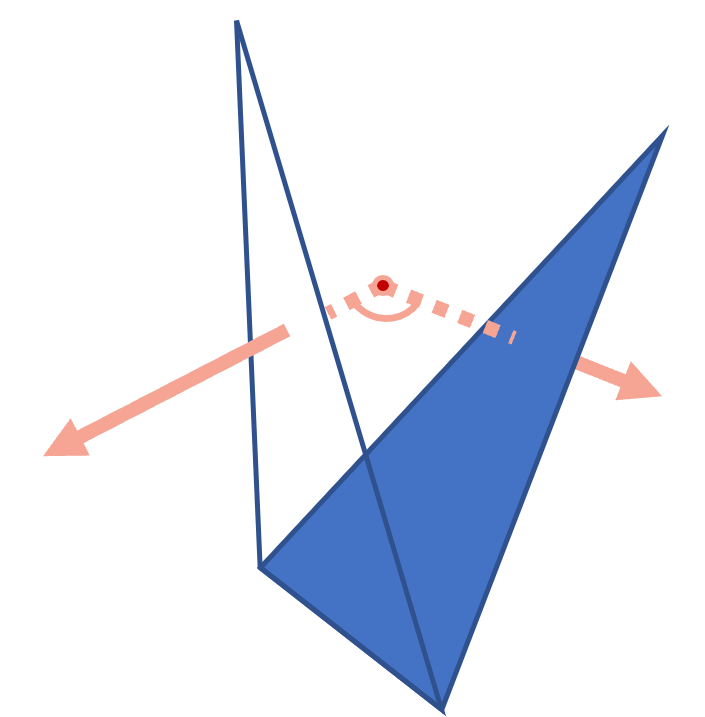}
        \caption{Easy Non-manifold Face}
        \label{fig:enmf}
    \end{subfigure}
    \hfill
    \begin{subfigure}[b]{0.49\columnwidth}
        \centering
        \includegraphics[height=0.6\columnwidth]{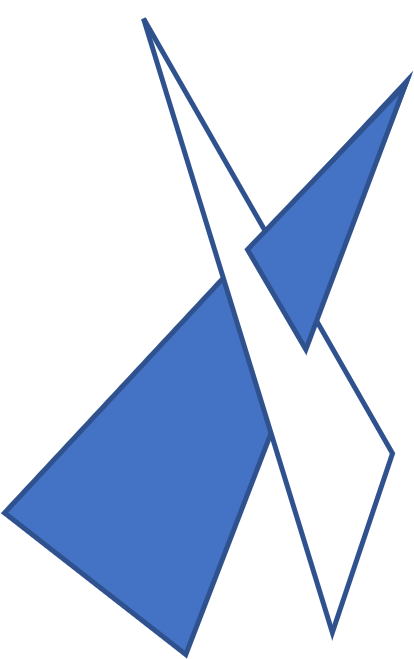}
        \caption{Self-intersection}
        \label{fig:self-intersection}
    \end{subfigure}
    \caption{\textbf{Unpleasant Faces} - (a): one face is a E-NMF if its normal direction is significant different from that of its adjacent faces; (b): SI faces cross other faces in the same mesh.}
    \label{fig:unpleasant-faces}
\end{figure}

We have described how we realize shape registration on top of dense point correspondence in Sec.~\ref{sec:pc and sr}. In our experiments of shape registration, the source meshes $\mathcal{M}_s$ are basically the template meshes and the target meshes $\mathcal{M}_t$ are all shape instances. 
To make different methods comparable, we apply Approximated Centroidal Voronoi Diagrams (ACVD) \cite{valette2004approximated, valette2008generic} to the template meshes of DIT, DIF-Net and NDF to make their template meshes meet the same resolution and similar topology. Specifically, we re-mesh these template meshes into $\mathcal{M}_s$ with 2500 vertices (clusters) or 5000 vertices. 

Apart from CD and NC, we design two more metrics to evaluate the geometrical fidelity of registration results: easy non-manifold face (E-NMF) ratio and self-intersection (SI) ratio, as shown in Fig.~\ref{fig:unpleasant-faces}, NMFs are such faces that have opposite normal direction to their adjacent faces. But in our scenario, this definition is too harsh because for organs such as lung and liver, the sudden change of face normal directions might take place in some local regions. As a result, one face will be defined as an E-NMF if the cosine similarity between normal directions of any of its adjacent faces and itself is less than $\delta$. It is set to be 0 when we evaluate heart and pancreas organs, and set to be -0.5 and -0.8 for lung and liver respectively. 

\begin{table*}
\centering
\resizebox{\linewidth}{!}{%
\begin{tabular}{clcccccccccccccccc} 
\toprule
 \# of           &         & \multicolumn{4}{c}{CD Mean($\downarrow$)} & \multicolumn{4}{c}{NC Mean($\uparrow$)} & \multicolumn{4}{c}{E-NMF Ratio Mean($\downarrow$)}               & \multicolumn{4}{c}{SI Ratio Mean($\downarrow$)}    \\ 
\cmidrule(r){3-6}\cmidrule(r){7-10}\cmidrule(r){11-14}\cmidrule(r){15-18}
Vertices              & Model /\ Organ   & Pancreas & Liver    & Lung     & Heart    & Pancreas & Liver    & Lung     & Heart    & Pancreas & Liver    & Lung     & Heart           & Pancreas & Liver    & Lung     & Heart   \\ 
\midrule
\multirow{5}{*}{2500} 
                      & AtlasNet\_Sph & 8.08  & 3.46  & 5.01  & 7.55   & 0.703    & 0.823 & 0.824 & 0.808           & 31     & \textbf{0.391} & \textbf{1.65}  & \textbf{0}    & 5860     & 29.5   & 13.8   & \textbf{0}       \\
                      & AtlasNet\_25  & 6.05  & 2.48  & 207   & 4.86   & 0.65     & 0.818 & 0.772 & 0.824           & 48.4   & 42.4  & 43.3  & 73.4 & 24500    & 25100  & 18700  & 24800   \\
                      & DIF-Net       & 7.44  & 1.74  & 1.98  & 2.44   & 0.736    & 0.834 & 0.879 & 0.842           & 540    & 16.8   & 63.9  & 76.1 & 4990     & 595      & 553    & 642  \\
                      & DIT           & 0.682 & 0.543 & 0.758 & 1.09   & 0.893    & 0.867 & 0.928 & 0.917           & 24.4   & 2     & 13.6  & 27.6 & 149      & 6.23   & 1.76   & \textbf{0}       \\
\cmidrule(r){2-18}
                      & \textbf{Ours}          & \textbf{0.53}  & \textbf{0.507} & \textbf{0.704} & \textbf{1.06}   & \textbf{0.915}    & \textbf{0.872} & \textbf{0.935} & \textbf{0.922}           & \textbf{0.191}  & 1.02  & 8.58  & 25.4 & \textbf{0}        & \textbf{0.89}   & \textbf{0}      & \textbf{0}       \\ 
\midrule
\multirow{3}{*}{5000} 
                      & DIF-Net           & 10.5  & 2.06  & 1.94  & 2.42   & 0.694    & 0.832 & 0.881 & 0.838           & 276    & 8.27  & 45.2  & 98.1 & 2560     & 4.61   & 786    & 1090  \\
                      & DIT           & 0.677 & 0.528 & 0.736 & 1.07   & 0.893    & 0.868 & 0.931 & 0.918           & 66.6   & 2.18  & 7.06  & 14.8 & 346      & 11.8   & 2.06   & \textbf{0}       \\
\cmidrule(r){2-18}
                      & \textbf{Ours}          & \textbf{0.518} & \textbf{0.49}  & \textbf{0.67}  & \textbf{1.02}   & \textbf{0.916}    & \textbf{0.873} & \textbf{0.936} & \textbf{0.923}           & \textbf{0.191}  & \textbf{0.378} & \textbf{2.68}  & \textbf{14.1} & \textbf{0}        & \textbf{2}      & \textbf{0}      & \textbf{0}       \\
\bottomrule
\end{tabular}
}
\caption{
\textbf{Shape Registration on Unseen Shape Instances} -- We align all unseen shape instance to the source mesh $\mathcal{M}_s$ for four organ categories. For AtlasNet, $\mathcal{M}_s$ is defined as their explicit template mesh. For DIF-Net, DIT and our NDF, $\mathcal{M}_s$ is defined as the re-meshed learned template mesh. We compare AtlasNet\_Sph and AtlasNet\_25 to the other methods having 2500 vetices in $\mathcal{M}_s$ after re-mesh. We also compare DIF-Net, DIT and NDF provided $\mathcal{M}_s$ is of 5000 vertices after remesh. Even though the number of vertices in $\mathcal{M}_s$ of AtlasNEt\_Sph is not 2500 (very close to), we still put them in comparison with methods having 2500-vertice $\mathcal{M}_s$. The E-NMF ratio results shown above are multiplied by $10^5$
} 
\label{tab:registration_test}
\end{table*}
Tab.~\ref{tab:registration_test} strongly supports that our NDF can densely align points across shapes while maintaining the topology. NDF achieves the best results in accuracy whatever experiment settings and organ classes are. Also, as for E-NMF ratio and SI ratio, our model can also outperform the other methods in most cases. AtlasNet\_Sph beats us on liver, lung and heart regarding E-NMF ratio in a price of the over-smoothened reconstruction results. Notably, in comparison with DIF-Net and DIT that are very competitive in shape representation and reconstruction, our method obviously outperforms them in all metrics due to the properties of deep diffeomorphic flow (Sec.~\ref{sec:ddit}). In summary, NDF is superior to the other state-of-the-art methods with respect to shape registration accuracy as well as fidelity by a great margin. We also report shape registration results on seen shape objects in supplementary material.

\subsection{Qualitative Results}
\begin{figure*}
    \centering
    \begin{subfigure}[b]{0.6\columnwidth}
        \centering
        \includegraphics[height=0.6\columnwidth]{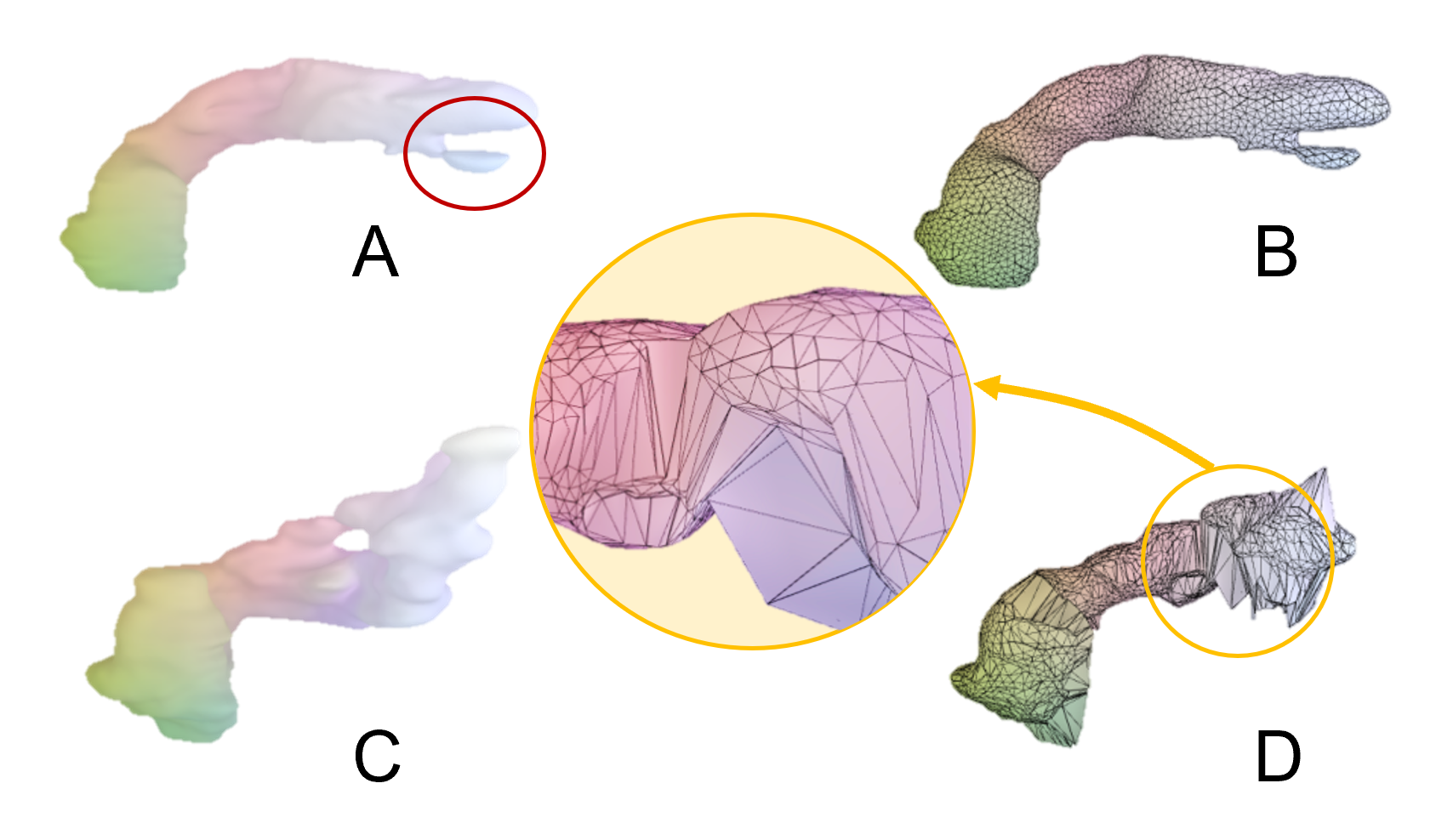}
        \caption{DIF-Net}
        \label{fig:dif_qualitative}
    \end{subfigure}
    \hfill
    \begin{subfigure}[b]{0.6\columnwidth}
        \centering
        \includegraphics[height=0.6\columnwidth]{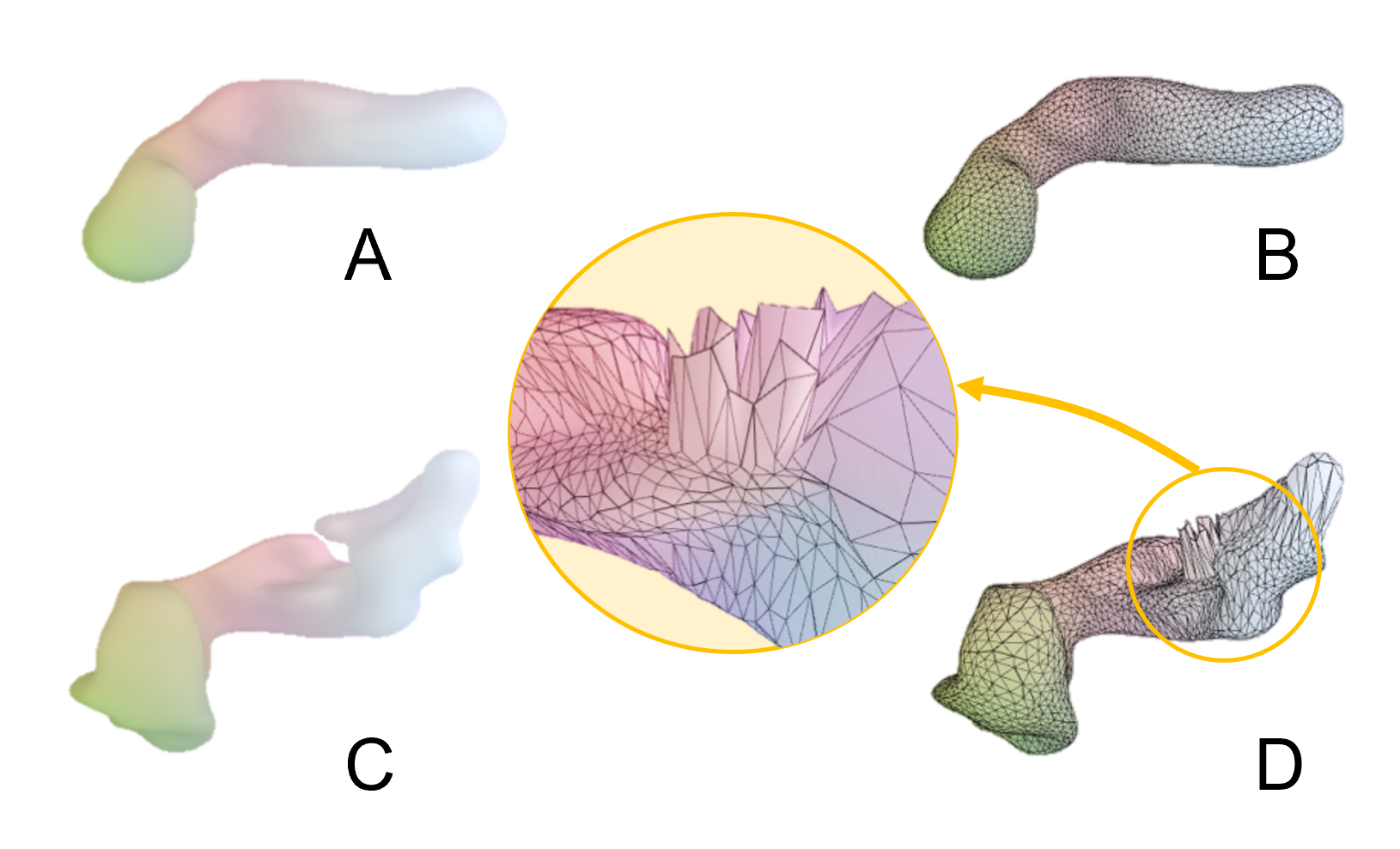}
        \caption{DIT}
        \label{fig:dit_qualitative}
    \end{subfigure}
    \hfill
    \begin{subfigure}[b]{0.6\columnwidth}
        \centering
        \includegraphics[height=0.6\columnwidth]{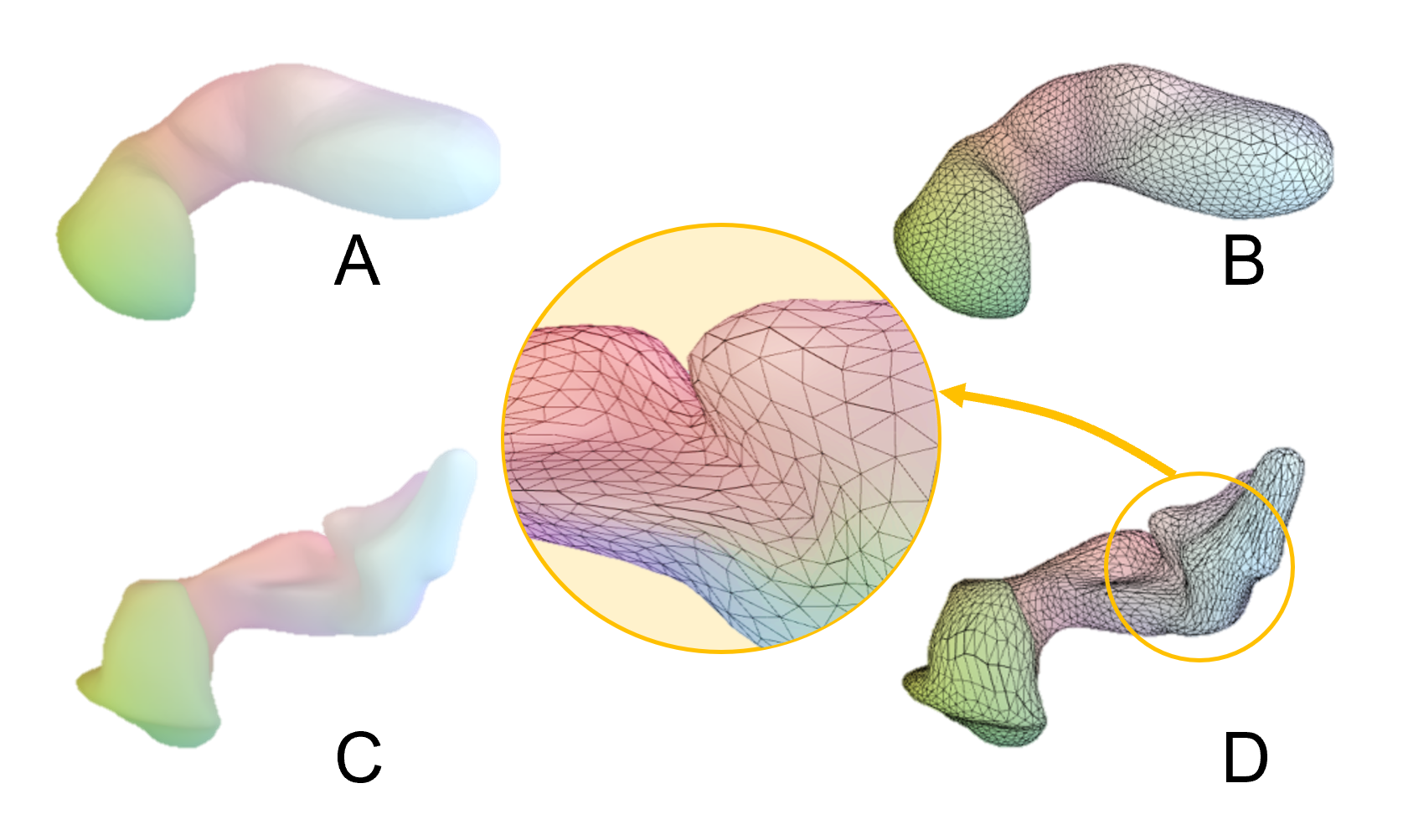}
        \caption{NDF}
        \label{fig:ddit_qualitative}
    \end{subfigure}
    \caption{\textbf{Qualitative Results Comparison} - For each method, they have four type of mesh shown above. Mesh A is the learned template mesh, Mesh B is the re-meshed template mesh with 2500 vertices, Mesh C is a reconstructed shape and Mesh D is the shape registration result. We also zoom in Mesh D for a better qualitative comparison.}
    \label{fig:qualitative_results}
\end{figure*}

Fig.~\ref{fig:teaser} demonstrates how NDF deforms shape instances to the learned templates in a coarse to fine manner while preserving the topology. Fig.~\ref{fig:qualitative_results} shows an example of pancreas shape representation and registration that can suggest why NDF stands out in Tab.~\ref{tab:registration_test}.. From Fig.~\ref{fig:dif_qualitative}, we can see the pancreas template learned by DIF-Net is problematic that the structure marked by a red circle has no anatomical meaning.
Different from DIF-Net and DIT which use nearest neighbours searching to match points, NDF is invertible and topology-preserved. Therefore, our shape registration results will have the comparable quality of shape reconstruction results. In Fig.~\ref{fig:dif_qualitative} and Fig.~\ref{fig:dit_qualitative}, there are many unpleasant triangles Fig.~(\ref{fig:unpleasant-faces}) in the local regions where the shape distortions between the shape instance and template are large. On the contrary, the registration result generated by NDF is clean, smooth and accurate. More qualitative results can be found in supplementary material.

\subsection{Ablation Study}
\label{sec.ablation}
\begin{table}
\centering
\resizebox{0.7\linewidth}{!}{%
\begin{tabular}{llcc} 
\toprule[.5pt]
  & Exp.                     & CD Mean   & NC Mean          \\ 
\midrule[.4pt]
1 & SV                       & 1.42      & 0.907   \\
2 & QTV$_4$                   & 0.516     & 0.914    \\
3 & QTV$_4$ + CL              & 0.51     & 0.916    \\
4 & QTV$_8$ + CL              & 0.51     & 0.916     \\
5 & \textbf{QTV$_4$ + CL + w/o pp}     & 0.512    & 0.917    \\
6 & TV                      & 1.65     & 0.902            \\
7 & TV + CL                 & 125      & 0.885            \\
\bottomrule[.5pt]
\end{tabular}
}
\caption{
\textbf{Ablation Study in Shape Reconstruction} -- The notations of experiments are described in Sec. \ref{sec.ablation}. The subscript of QTV is the number of progressive reconstruction steps. These experiments are conducted on the unseen pancreas shapes.
} 
\label{tab:ablations_rec_test}
\end{table}

\begin{table}
\centering
\resizebox{\linewidth}{!}{%
\begin{tabular}{llcccc} 
\toprule
  & Exp.                 & CD Mean & NC Mean & E-NMF Mean & SI Mean \\ 
\midrule
1 & SV                   & 1.46    & 0.906   & 1.43     & 0        \\
2 & QTV$_4$               & 0.521   & 0.914   & 0        & 0        \\
3 & QTV$_4$ + CL          & 0.52    & 0.915   & 0        & 0        \\
4 & QTV$_8$ + CL          & 0.521   & 0.915   & 2.38     & 0        \\
5 & \textbf{QTV$_4$ + CL + w/o pp} & 0.518   & 0.916   & 1.91     & 0        \\
6 & TV                  & 1.65    & 0.902   & 0.953    & 0        \\
7 & TV + CL             & 162     & 0.884   & 12.9     & 0.103    \\
\bottomrule
\end{tabular}
}
\caption{
\textbf{Ablation Study in Shape Registration} -- All these experiments are conducted on unseen pancreas shapes given the $\mathcal{M}_s$ with 5000 vertices.
} 
\label{tab:ablations_reg_test}
\end{table}

Our ablation study is developed on pancreas shapes to investigate the effects of three designs in our approach. In this section, "SV", "QTV", "TV" sequentially stands for stationary, our quasi time-varying and time-varying velocity field, "CL" is the short term for curriculum learning and "pp" denotes the point pair loss, which acts as a smooth regularization. Our final approach is labeled as Exp.5 in Tab.~\ref{tab:ablations_rec_test} and Tab.~\ref{tab:ablations_reg_test}. 

\paragraph{Quasi Time-varying Velocity Field}
From the comparisons among Exp.1, 2 and 6 in Tab.~\ref{tab:ablations_rec_test}, our quasi time-varying velocity field wins in all aspects. Time-varying velocity field should be a natural choice but without enough temporal information, the generability will be questionable. Stationary velocity field is widely applied in the medical image/surface registration problem, but it turns out to be hard to get the optimal if assume the continuous velocity field is independent of time. We also explore the effect of progressive representation steps by comparing Exp.4 and 8, we have not observed some extra improvements resulting from more representation steps. Fig.~\ref{fig:teaser} also indicates most deformations have been done in the starting phrases.

\paragraph{Curriculum learning}
The improvements earned from curriculum learning are significant in other work \cite{duan2020curriculum, zheng2021deep}. While in our work, it is beneficial but not essential, as can be seen in Tab.~\ref{tab:ablations_rec_test} and Tab.~\ref{tab:ablations_reg_test}. Furthermore, it is even harmful when used together with a time-varying velocity field. 

\paragraph{Smooth Regularization}
Our opinion that the smooth regularization term is not needed when training deep diffeomorphic flow is supported. In Tab.~\ref{tab:ablations_rec_test} and \ref{tab:ablations_reg_test}, we can see Exp.5 and Exp.6 get the very close reconstruction accuracy and both of them generate very few unpleasant faces in shape registration. In summary, even with the most basic training loss and training strategy, our design of model can get a very competitive performance in shape reconstruction and registration.
\section{Applications and Limitations}
\subsection{Applications}
NDF keeps most of the benefits of DeepSDF such as shape completion and shape interpolation. As for medical meaning, NDF can help post-process the segmentation results and do plausible data augmentation.

\begin{figure}[t]
\begin{center}
\includegraphics[width=\columnwidth]{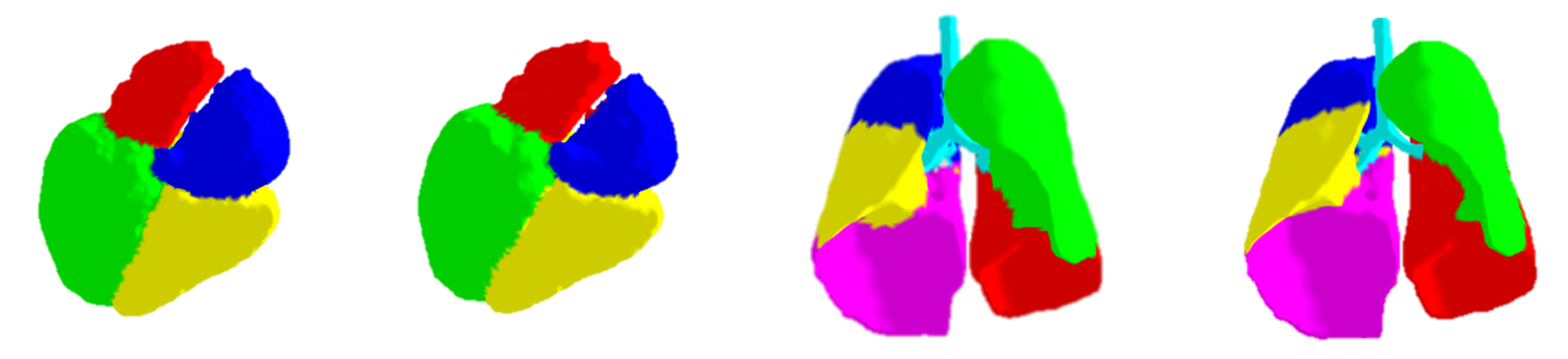}
\end{center}
\caption{\textbf{Label Transfer} - \textit{LeftMost}: Sub-heart ground truth; \textit{MiddleLeft}: Label Transfer; \textit{MiddleRight}: Lobe groud truth; \textit{RightMost}: Label Transfer}
\label{fig:label_transfer}
\end{figure}
Our model can also help transfer labels from seen shape to unseen shape. We choose 5 samples from the training set and transfer their labels to the target shape separately. The final label is the majority voting results. Fig.~\ref{fig:label_transfer} shows two examples of labels transferred by NDF.

The ambition of our work is to boost the shape analysis in medical imaging by helping establish organ shape dataset having dense topology-preserving point correspondences. Specifically, as long as our model is trained on one class of organ shapes and implicit template mesh $\mathcal{M}_T = (\mathcal{V}_T, \mathcal{E}_T)$ is labelled, the organ mesh of the same class could be aligned as Sec.~\ref{sec:pc and sr} explains. Given such dataset, we can learn a model to parameterize shapes as SMPL \cite{loper2015smpl}.

\subsection{Limitations}

\begin{figure}[t]
\begin{center}
\includegraphics[width=\columnwidth]{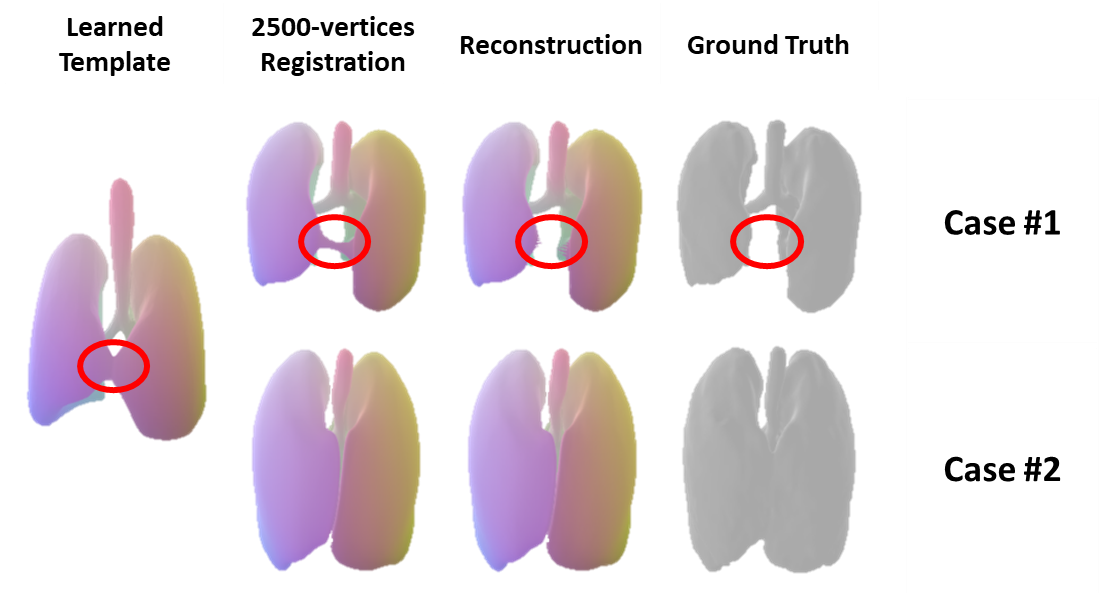}
\end{center}
\caption{\textbf{Limitations} - Case \#1 has different local structure with learned template. Case \#2 is a successful example.}
\label{fig:limitations}
\end{figure}
Our model concentrates on reconstructing and matching a group of shapes sharing common structures, so we haven't applied it to the popular 3D shape datasets like ShapeNet \cite{chang2015shapenet}. 
%
%
As can be seen in Fig.~\ref{fig:limitations}, our learned template has structures (marked by red circle) that don't exist in case \#1, then our shape reconstruction and registration results of case \#1 shape is negatively affected by the extra structures. This issue can be partially addressed by introducing the correction module \cite{deng2021deformed} or considering shapes as groups of sub-structures \cite{liu2020learning} that are individually topology-preserving or non-existent. 

To our best knowledge, there is no medical dataset having structures as well as dense point correspondences annotated. Thus, we cannot evaluate shape registration results in terms of point-to-point error. In the future, we will explore the potentials of our model on some synthetic data like D-FAUST \cite{bogo2017dynamic}, with which we can do point-to-point analysis. We will present  The inference runtime of our method is indeed longer than competing methods such as DiT. 
The main bottleneck of our method is the neural ODE (NODE) module, which requires repeated functional evaluation to solve ODEs within a given error tolerance. There are two potential solutions to shorten the inference time: 1) as \cite{IshitMehta2021ModulatedPA}, apply modulation network maps a latent code to the modification of parameters of a base network.; 2) as NPMs \cite{palafox2021npms}, train an encoder separately to overfit the learned latent code and utilize this encoder to generate the initialization for latent code optimization.

\section{Conclusions}

In this paper, we propose a novel deep implicit function based on neural diffeomorphic flow (NDF) for topology-preserving shape representation. Our experimental results demonstrate that
explicitly considering topology preservation leads to significant improvements on shape representation and registration, as illustrated on medical images, where topology preservation is often a necessary requirement. 
We also propose a conditional quasi time-varying approach to model NDF through an auto-decoder model consisting of multiple neural ODE blocks, allowing us to model the shape deformation in a progressive manner. 
%

{
    \small
    \bibliographystyle{ieee_fullname}
    \bibliography{macros,main}
}

\appendix

\setcounter{page}{1}

\twocolumn[
\centering
\Large
\textbf{Supplementary Material} \\
\vspace{1.0em}
] 

\appendix

\section{Implementation Details}
\subsection{Deformation Module}

\begin{figure}[!ht]
\begin{center}
\includegraphics[width=\columnwidth]{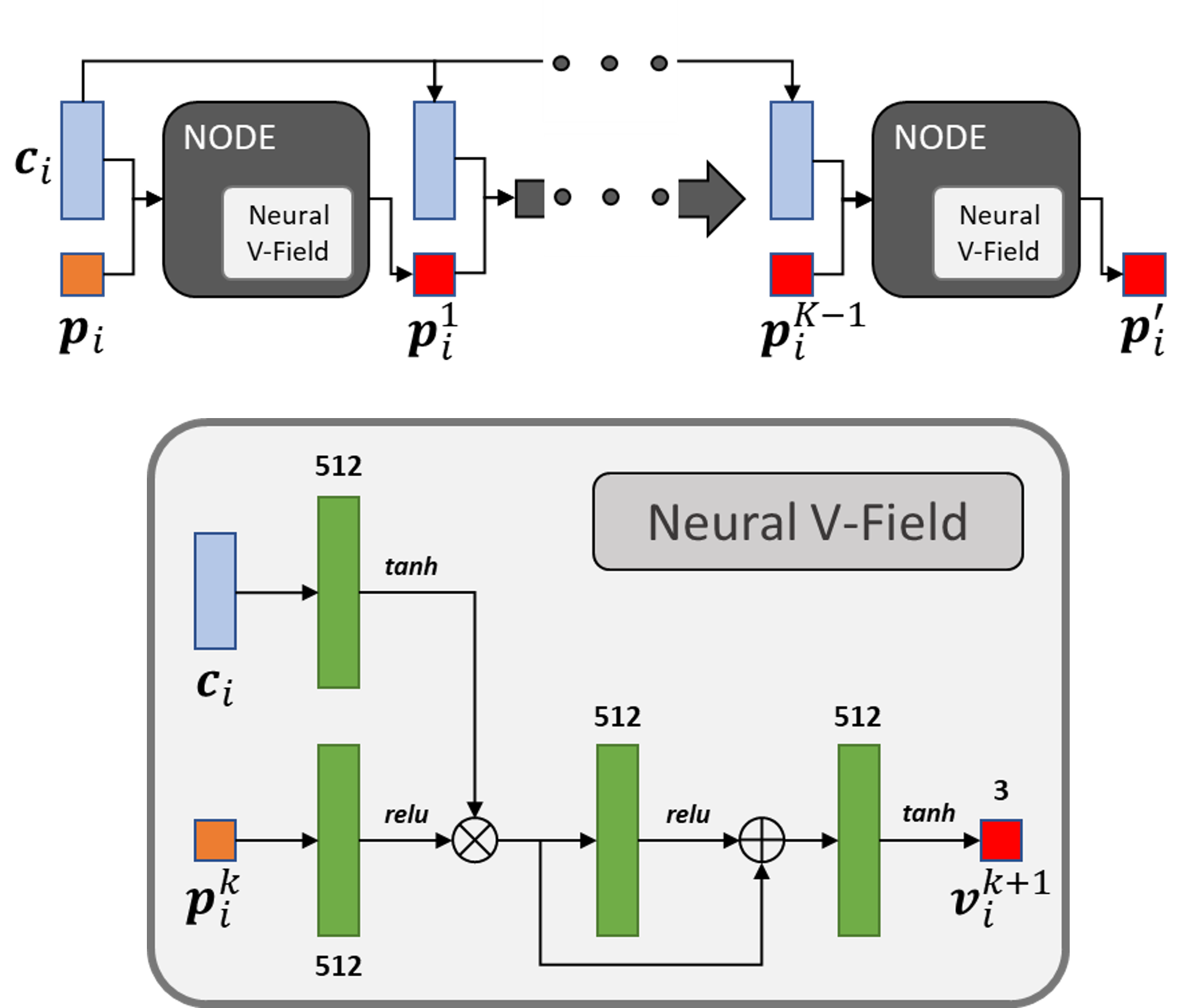}
\end{center}
\caption{Deformation Module}
\label{fig:ndf}
\end{figure}
In implementation, we realize the deformation module with several concatenated NODE blocks with neural velocity field as the dynamics function. As shown in Fig.~\ref{fig:ndf}, the integral of the last NODE block, which is the intermediate deformed position, will be the input of the next NODE block, together with the unchanged latent code of the shape. $\boldsymbol{p}_i$ are the original points in the SDF of shape $X_i$, $\boldsymbol{p}_{i}^{k}$ is the deformed positions of $\boldsymbol{p}_i$ after the $k$-th NODE block and $\boldsymbol{p}_{i}^{'}$ is the final deformed position of $\boldsymbol{p}_i$ in the template space.

The neural velocity field is simply residual blocks based on fully connected layers, which are represented as green blocks in Fig.~\ref{fig:ndf}. The hidden feature size are set to be 512 and the final activate function is set to be \textit{tanh} because we want the deformed positions are within the normalized range [-1, 1]. 
In Fig.~\ref{fig:ndf}, we multiply the point features and shape features, but it is not necessary. There should be no significant difference if the point features and shape features are concatenated instead.

\subsection{Training and Inference Setting}
During training and inference, the tolerance of NODE solver is set as $1e^{-5}$. We set the length of deform code as 256 and the points sampled from each shape objects are 8000, of which 4000 are inside points and the others are outside points. In inference, the points sampled for optimizing deform code could be partial or complete. For each organ class, We train a NDF for 2000 epochs using Adam \cite{kingma2014adam} with a learning rate $5e-4$ and batch size of 8. 
Also, we jointly optimize the deform codes along with NDF training using Adam with a learning rate $1e-3$. 
The hyper-parameters $\epsilon$ and $\lambda$ follows the setting of \cite{zheng2021deep} and regularization loss weight $\lambda_{p w}$ is set to be 1. During inference, the deform codes is optimized for 2400 iterations with a learning rate $5e-2$.

DIT uses the exactly same training and inference setting as NDF. We train DeepSDF, DIF and AtlasNet for all organs with the default settings for airline shapes provided by their official implementations.  

\section{Data Source and Data Preparation}
Pancreas CT dataset contains 82 abdominal contrast enhanced 3D CT scans with pancreas labelled. 
We split 61 samples for training and 21 for testing. 

Multi-Modality Whole Heart Segmentation (MMWHS) challenge \cite{zhuang2016multi} provided 60 labelled multi-modality medical images with 7 whole heart substructure annotated, including left and right ventricle blood cavity, left and right atrium blood cavity, myocardium of left ventricle, ascending aorta as well pulmonary artery. In our experiments, we construct the whole heart label with the union of the annotations of blood cavity and myocardium of left ventricle. The ratio between training set and testing set is $3:1$.

Lung dataset combines the 85 chest data from \cite{chen2021deep} and 51 lobe segmentation data from \cite{tang2019automatic}. The six sub-structures labelled by lobe segmentation data are upper left lung, middle left lung, lower middle lung, upper right lung, lower right lung and trachea. The organ shape we learn to reconstruct is the union of lung and trachea for both chest data ans lobbe segmentation data. We put 85 chest data and 17 lobe segmentation data into training set and 34 lobe segmentation data in testing set. 

Liver dataset collects 190 samples coming from \cite{chen2021deep} with liver annotations. We split them into 145 and 45 shape instances into training and testing set respectively.

Since all these organs are labelled as 3D volumes, we first extract the organ surface using Marching Cube, then follow \cite{park2019deepsdf} to sample normalized SDF points near the mesh surface.

\section{Experiments}

\subsection{Seen Shape Representation}
\begin{table*}
\centering
\resizebox{\linewidth}{!}{%
\begin{tabular}{lcccccccccccccccc} 
\toprule
               & \multicolumn{4}{c}{CD Mean ($\downarrow$)} & \multicolumn{4}{c}{CD Median ($\downarrow$)} & \multicolumn{4}{c}{NC Mean ($\uparrow$)} & \multicolumn{4}{c}{NC Median ($\uparrow$)}  \\ 
\cmidrule(lr){2-5}\cmidrule(lr){6-9}\cmidrule(lr){10-13}\cmidrule(lr){14-17}
Model / Organ         & Pancreas & Liver    & Lung     & Heart    & Pancreas & Liver    & Lung     & Heart    & Pancreas & Liver    & Lung     & Heart           & Pancreas & Liver    & Lung     & Heart              \\ 
\midrule
AtlasNet\_Sph & 4.5   & 1.76  & 3.64  & 5.03  & 4.08  & 1.39  & 3.21  & 4.64  & 0.733    & 0.836 & 0.82  & 0.817           & 0.736    & 0.841 & 0.824 & 0.822              \\
AltlasNet\_25 & 5.48  & 1.9   & 8.97  & 3.08  & 3.06  & 0.985 & 1.86  & 2.32  & 0.674    & 0.833 & 0.828 & 0.827           & 0.684    & 0.835 & 0.837 & 0.83               \\
DeepSDF        & \underline{0.34}  & \underline{0.232} & \underline{0.247} & \underline{0.375} & 0.335 & \underline{0.226} & \underline{0.244} & \underline{0.359} & 0.927    & 0.876 & \underline{0.933} & 0.936           & 0.931    & 0.877 & 0.933 & 0.94               \\
DIF-Net            & 0.568 & \textbf{0.122} & \textbf{0.122} & \textbf{0.243} & \textbf{0.205} & \textbf{0.102} & \textbf{0.118} & \textbf{0.245} & \textbf{0.979}    & \textbf{0.894} & 0.856 & \textbf{0.961}           & \textbf{0.981}    & \textbf{0.895} & 0.856 & \textbf{0.965}              \\
DIT            & 0.349 & 0.303 & 0.682 & 0.632 & 0.343 & 0.287 & 0.376 & 0.583 & 0.929    & 0.878 & 0.931 & 0.934           & 0.935    & 0.878 & \underline{0.934} & 0.94               \\
\midrule
\textbf{Ours}           & \textbf{0.315} & 0.291 & 0.351 & 0.479 & \underline{0.309} & 0.281 & 0.343 & 0.45  & \underline{0.933}    & \underline{0.883} & \textbf{0.939} & \underline{0.944}           & \underline{0.937}    & \underline{0.884} & \textbf{0.94}  & \underline{0.949}              \\
\bottomrule
\end{tabular}
}
\caption{
\textbf{Shape Representation} -- We demonstrate the reconstruction results of different representation methods on four organ categories. 
AtlasNet\_Sph and AtlasNet\_25 are AtlasNet using 3D sphere mesh and 25 square patches as the template shape respectively. Lower is better for chamfer distance ($\times 10^3$) and worse for normal consistency. Bold numbers are the best and the underlined are the second best.
} 
\label{tab:representation}
\end{table*}
From Tab.~\ref{tab:representation}, we can see DIF-Net \cite{deng2021deformed} performs best in terms of both CD and NC. As we have discussed in the main paper, we believe it is because of the extra points sampled on the shape surface, which could provide you more accurate representation about the zero level set of SDF. But as shown in main paper, the reconstruction performance of DIF-Net is not competitive compared to other deep implicit functions like DeepSDF \cite{park2019deepsdf} and DIT \cite{zheng2021deep}. Involving many surface points in training will make deep implicit functions too "explicit" to handle the noise in the raw data. In general, DeepSDF performs better than our work and DIT on seen shape representation in terms of CD. It is because, DeepSDF has latent code with more freedom that it is said to represent one shape. But in DIT and NDF, the latent code controls the deformations of one shape. Thus, DeepSDF is more likely to train and overfit but relies less on the understanding of shape priors. Therefore, under a small number of training samples, we can witness DeepSDF is inferior to DIT and NDF on unseen shape reconstruction. What's more, compared to DIT, our model is better in all organs and both evaluation metrics. 

\subsection{Seen Shape Registration}
\begin{table*}
\centering
\resizebox{\linewidth}{!}{%
\begin{tabular}{clcccccccccccccccc} 
\toprule
 \# of           &         & \multicolumn{4}{c}{CD Mean} & \multicolumn{4}{c}{NC Mean} & \multicolumn{4}{c}{E-NMF Ratio Mean}               & \multicolumn{4}{c}{SI Ratio Mean}    \\ 
\cmidrule(r){3-6}\cmidrule(r){7-10}\cmidrule(r){11-14}\cmidrule(r){15-18}
Vertices              & Model /\ Organ   & Pancreas & Liver    & Lung     & Heart    & Pancreas & Liver    & Lung     & Heart    & Pancreas & Liver    & Lung     & Heart           & Pancreas & Liver    & Lung     & Heart   \\ 
\midrule
\multirow{5}{*}{2500} 
                      & AtlasNet\_Sph & 4.5   & 1.76  & 3.64  & 5.03   & 0.733    & 0.836 & 0.82  & 0.817           & 26.7   & 1.13  & \textbf{1.36}  & \textbf{0}    & 5820     & 289    & 19.8   & \textbf{0}       \\
                      & AtlasNet\_25  & 5.48  & 1.9   & 8.97  & 3.08   & 0.674    & 0.833 & 0.828 & 0.827           & 55.3   & 36.9  & 68.2  & 71.2 & 24700    & 25600  & 24800  & 26300   \\
                      & DIF-Net           & 3.69  & 0.584 & \textbf{0.372} & 1.08   & 0.847    & 0.868 & 0.926 & 0.893           & 407    & 77.6  & 77.1  & 107  & 6870     & 32.6   & 933    & 1450  \\
                      & DIT           & 0.377 & 0.312 & 0.848 & 0.678  & 0.92     & 0.875 & 0.922 & 0.931           & 11.4   & 1.41  & 20.6  & 25.7 & 5.91     & 9.45   & 410    & \textbf{0}       \\
                      & \textbf{Ours}          & \textbf{0.326} & \textbf{0.309} & 0.406 & \textbf{0.528}  & \textbf{0.93}     & \textbf{0.881} & \textbf{0.934} & \textbf{0.941}           & \textbf{0.492}  & \textbf{0.221} & 19.8  & 30.4 & \textbf{0.656}    & \textbf{0.276}  & \textbf{15.9}   & \textbf{0}       \\ 
\midrule
\multirow{3}{*}{5000} 
                      & DIF-Net           & 3.65  & 0.561 & \textbf{0.345} & 1.02   & 0.849    & 0.871 & 0.933 & 0.894           & 435    & 16.2  & 54.9  & 121  & 6000     & 79.2   & 1380   & 1890  \\
                      & DIT           & 0.367 & 0.312 & 0.882 & 0.678  & 0.922    & 0.875 & 0.924 & 0.931           & 28.2   & 1.41  & 17.9  & 25.7 & 33.1     & 9.45   & 406    & \textbf{0}       \\
                      & \textbf{Ours}          & \textbf{0.32}  & \textbf{0.299} & 0.387 & \textbf{0.502}  & \textbf{0.932}    & \textbf{0.882} & \textbf{0.936} & \textbf{0.943}           & \textbf{0.246}  & \textbf{0.145} & \textbf{8}     & \textbf{15.6} & \textbf{0}        & \textbf{0}      & \textbf{9.21}   & \textbf{0}       \\
\bottomrule
\end{tabular}
}
\caption{
\textbf{Shape Registration on seen Shape Instances}
} 
\label{tab:registration_train}
\end{table*}
As the unseen shape registration results, our model can achieve the best performance with great advantages over the other methods. AtlasNet\_Sph performs good at the number of unpleasant faces and even better than NDF on Lung and heart in terms of E-NMF and SI ratio. But they are very poor in shape registration accuracy, which reveals that their shape reconstruction results are over-smoothened. Compared to DIT and DIF-Net, our results are mostly better, especially in terms of the unpleasant faces numbers. 

\subsection{Label Transfer}
Here, we will provided quantitative results of label transfer realized by DIT and NDF. 
We investigate the label transfer IOU performance on the sub-structures labelled in MMWHS dataset and lobe segmentation dataset. 
For this task, we first choose 5 source meshes from labelled training samples and apply point correspondence towards all test samples based on these 5 source meshes. Now we have the align points of each test shapes with labels. Lastly, we label each vertex of test mesh with the nearest labeled aligned points. We didn't compare the results of DIF and AtlasNet because they are not even close to the registration performance of our method. 

\begin{table}
\centering
\resizebox{0.7\linewidth}{!}{%
\begin{tabular}{llcc} 
\toprule[.5pt]
Model                    & MMWHS            & LOBE              & Mean      \\ 
\midrule[.4pt]
DIT                      & 89.94            & 76.63             & 83.12     \\
Ours                     & \textbf{92.38}   & \textbf{76.66}    & \textbf{84.52}     \\

\bottomrule[.5pt]
\end{tabular}
}
\caption{Label IOU on the label transfer task} 
\label{tab:label_transfer}
\end{table}

Tab.~\ref{tab:label_transfer} reflects that our method can out-perform DIT on both MMWHS and lobe segmentation datasets in the label transfer task. The high IOU score reveals 
that our work can distinguish the semantic sub-regions regardless of the shape variances and has the potential to be applied in few-shot 3D shape segmentation learning.

\subsection{Point-to-point Error}
We conducted an experiment on a motion sequence (chicken wings dance) from the D-FAUST dataset \cite{bogo2017dynamic}, containing dense point correspondence. 
We used the point-to-point euclidean distance to evaluate the registration accuracy
and compared our method to DiT (the previous state-of-the-art). Our method yields significantly better results in both accuracy (L2 distance) and quality (SI, self-intersection ratio) (Tab.~\ref{tab:d-faust_reg}). 
\begin{table}[h]
\centering
\footnotesize
\centering
\begin{tabular}{lcccc}
\toprule
\multicolumn{1}{l}{} & \multicolumn{2}{c}{train}         & \multicolumn{2}{c}{test}          \\ \cmidrule(lr){2-3}\cmidrule(lr){4-5}
Model       & L2 ($\downarrow$) & SI ($\downarrow$) & L2 ($\downarrow$) & SI ($\downarrow$) \\ \midrule
DiT                  & 0.0124          & 0.0646          & 0.0121          & 0.0990          \\
\textbf{ours}        & \textbf{0.0028} & \textbf{0.0215} & \textbf{0.0032} & \textbf{0.0222} \\ \bottomrule
\end{tabular}
\caption{
Registrations results on D-FASUT sequence
} 
\label{tab:d-faust_reg}
\end{table}

\subsection{Qualitative Results}
We encourage readers to zoom in the following illustrations for more detailed comparisons.

\paragraph{Neural Diffeomorphic Flow}
\begin{figure*}
\begin{center}
\includegraphics[width=\linewidth]{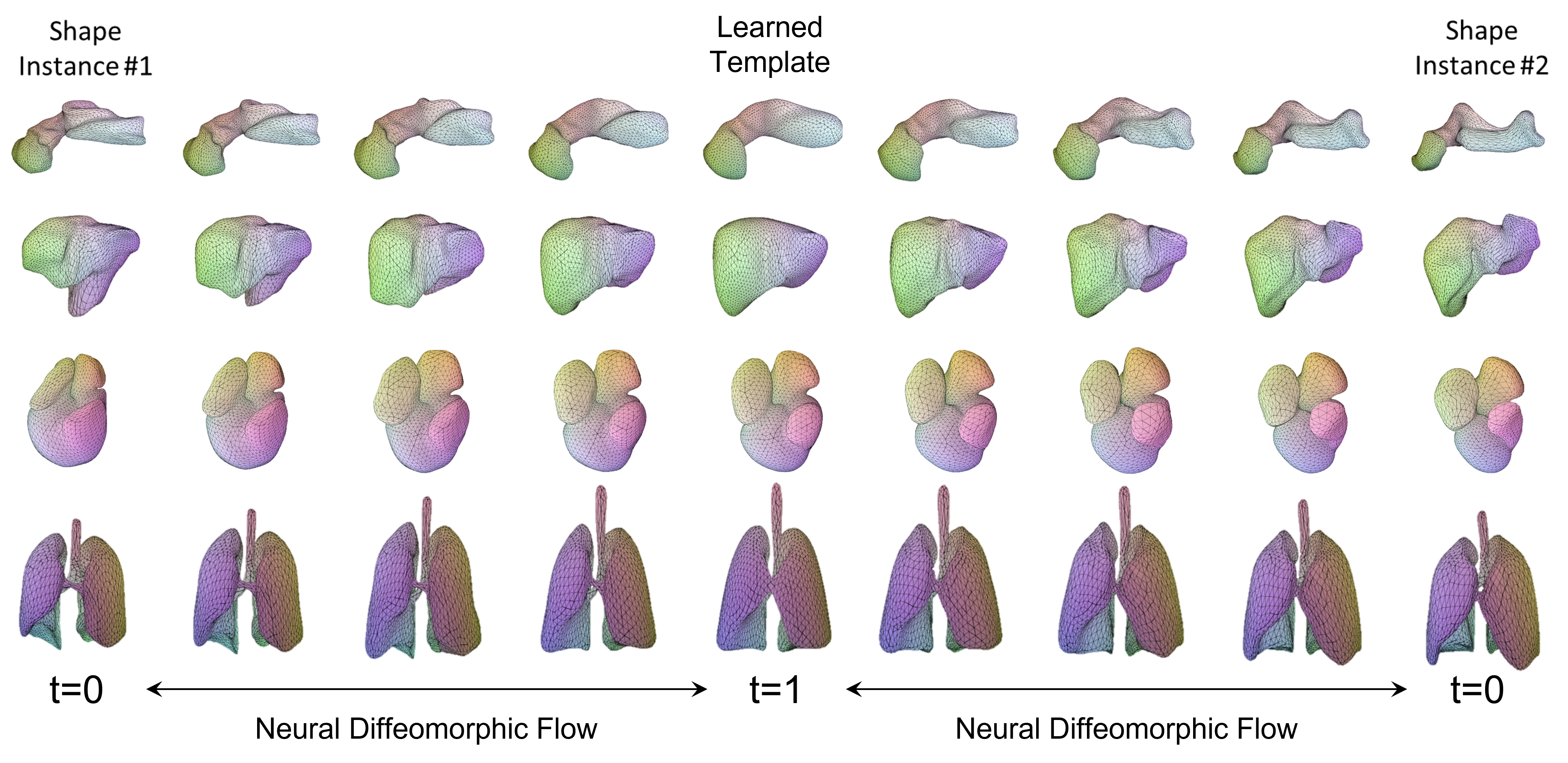}
\end{center}
\caption{Demonstration of our representation 
}
\label{fig:teaser_double}
\end{figure*}
Fig.~\ref{fig:teaser_double} demonstrate how two shapes find their correspondence via the learned template and show the intermediate results.

\paragraph{Template Shapes}
\begin{figure}
\begin{center}
\includegraphics[width=0.75\columnwidth]{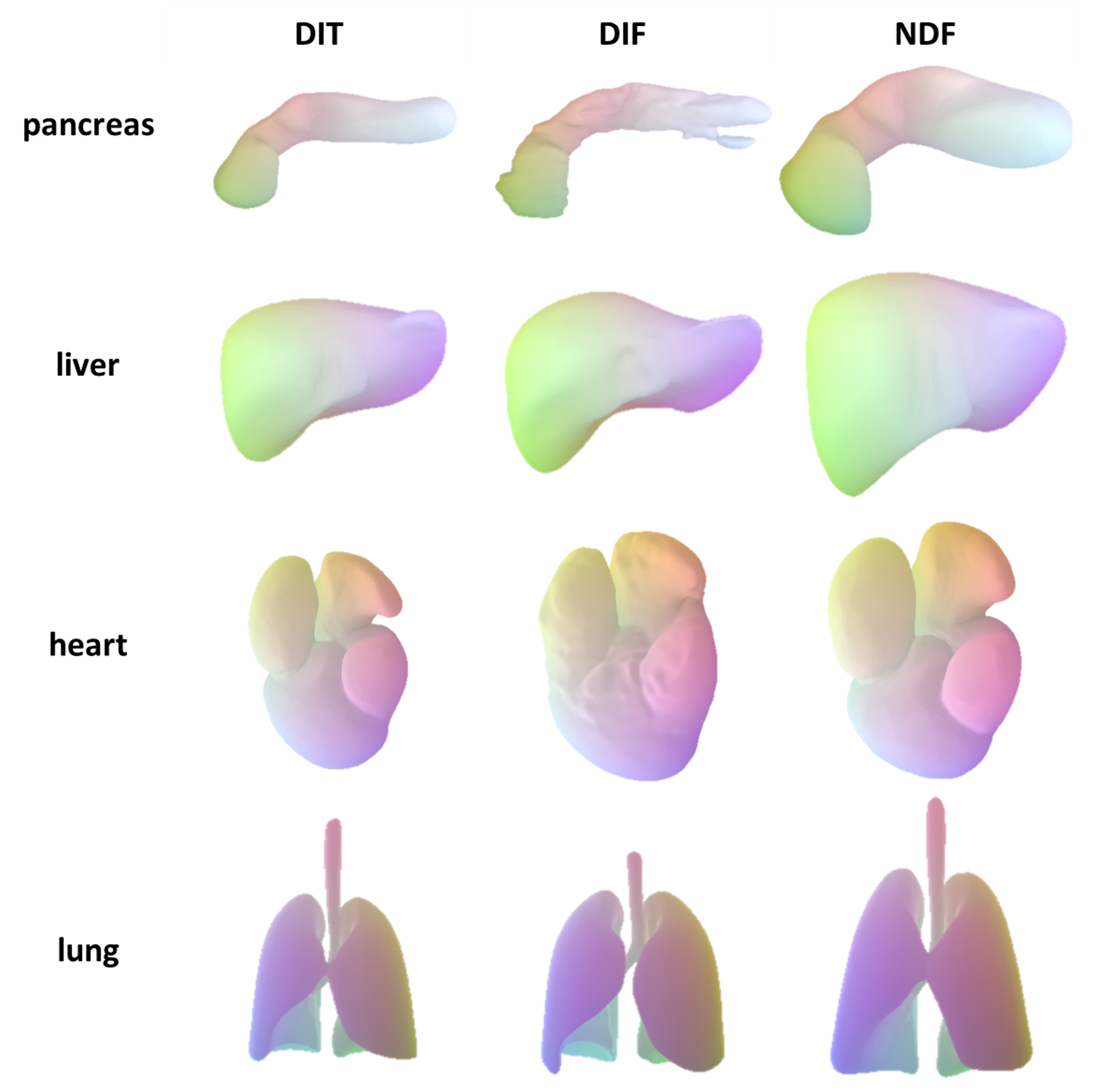}
\end{center}
\caption{Template shapes learned by DIF, DIT and NDF}
\label{fig:template_shapes}
\end{figure}
Fig.~\ref{fig:template_shapes} presents the template shapes learned by DIF, DIT and NDF. 

\paragraph{Seen/Unseen Shape Representation/Reconstruction}
\begin{figure*}
\centering
    \begin{subfigure}[b]{\columnwidth}
        \centering
        \includegraphics[width=\columnwidth]{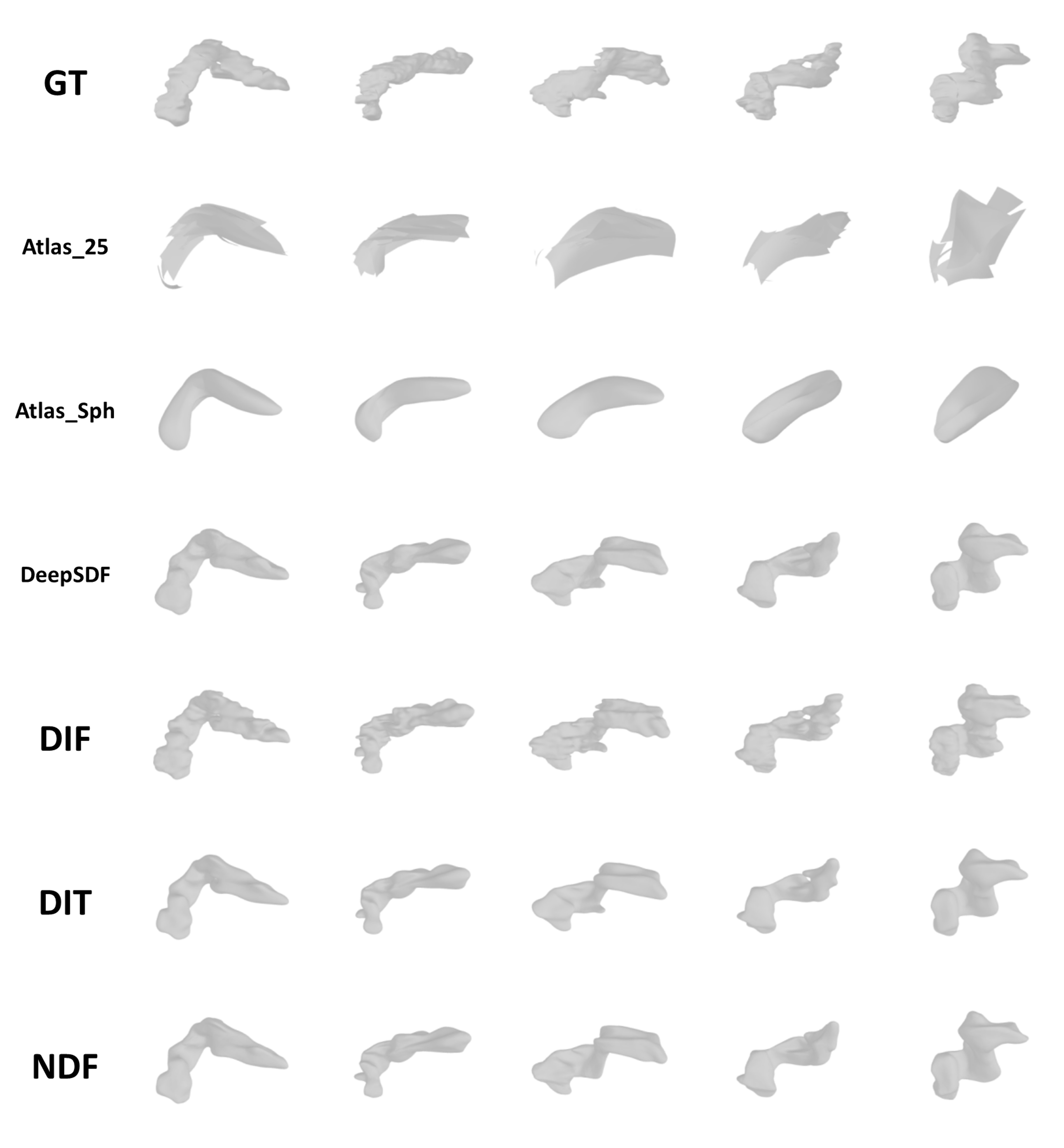}
        \caption{Pancreas}
        \label{fig:pancreas_train_rec}
    \end{subfigure}
    \hfill
    \begin{subfigure}[b]{\columnwidth}
        \centering
        \includegraphics[width=\columnwidth]{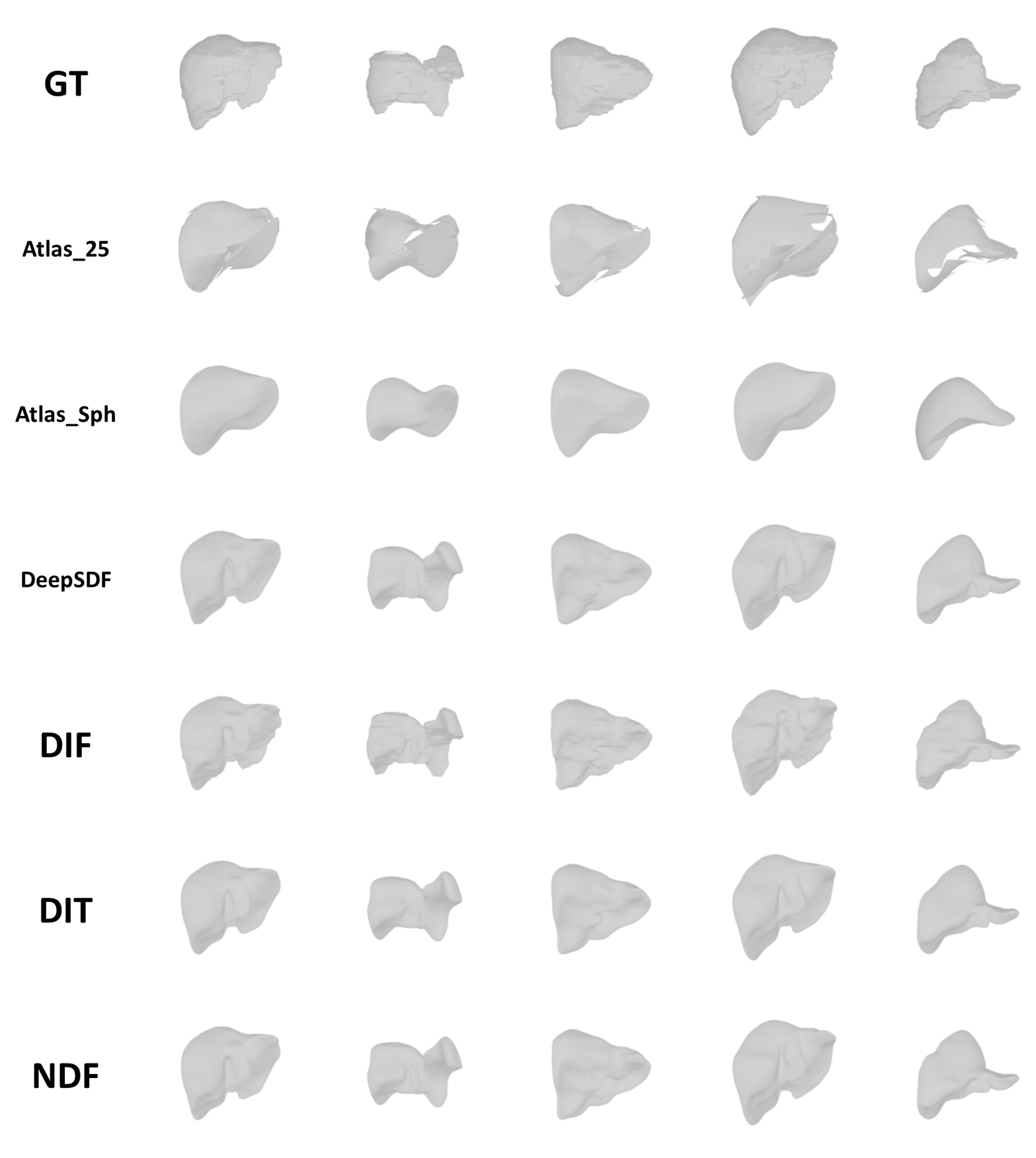}
        \caption{Liver}
        \label{fig:liver_train_rec}
    \end{subfigure}
    \vfill
    \begin{subfigure}[b]{\columnwidth}
        \centering
        \includegraphics[width=\columnwidth]{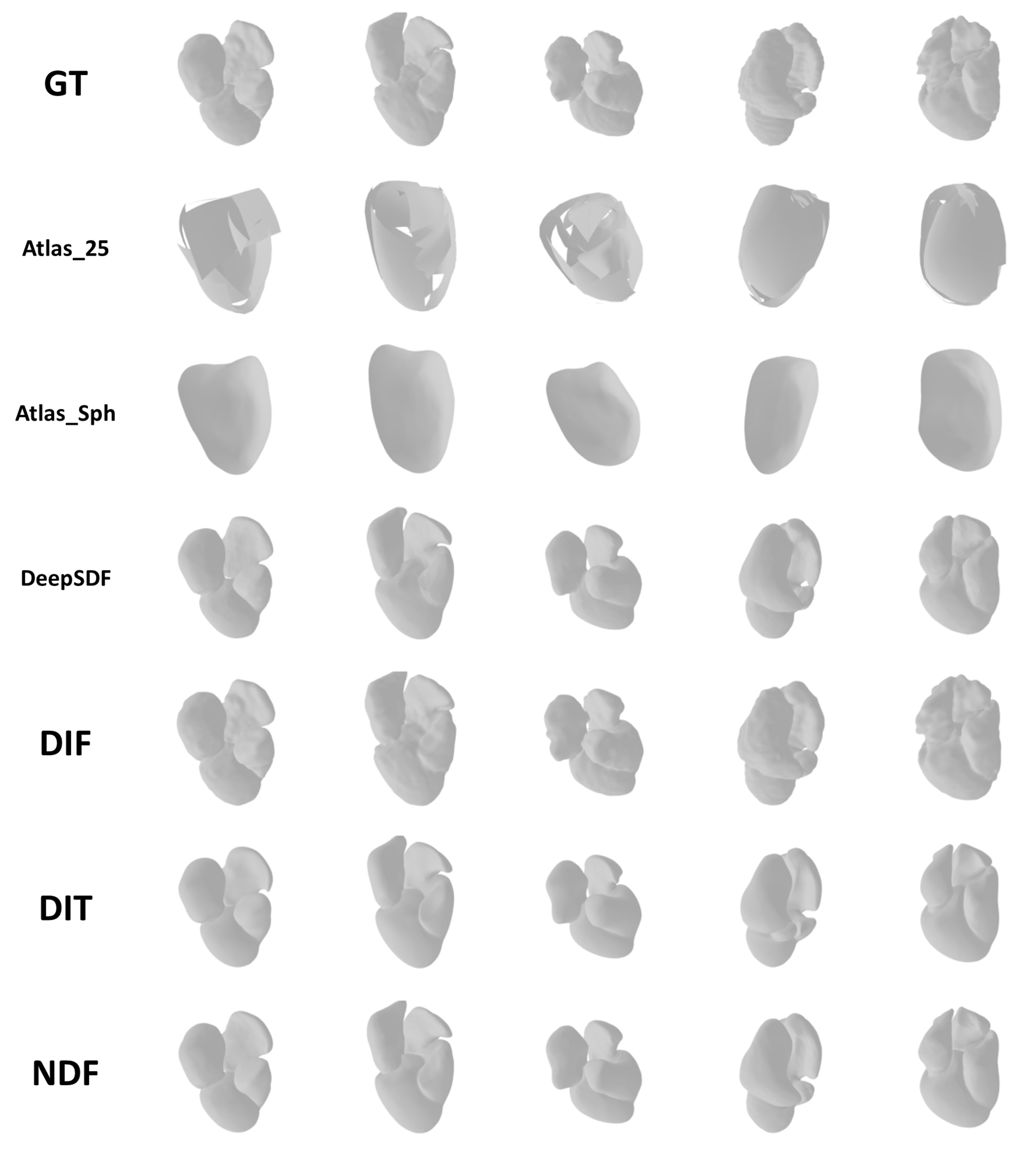}
        \caption{Heart}
        \label{fig:mmwhs_train_rec}
    \end{subfigure}
    \hfill
    \begin{subfigure}[b]{\columnwidth}
        \centering
        \includegraphics[width=\columnwidth]{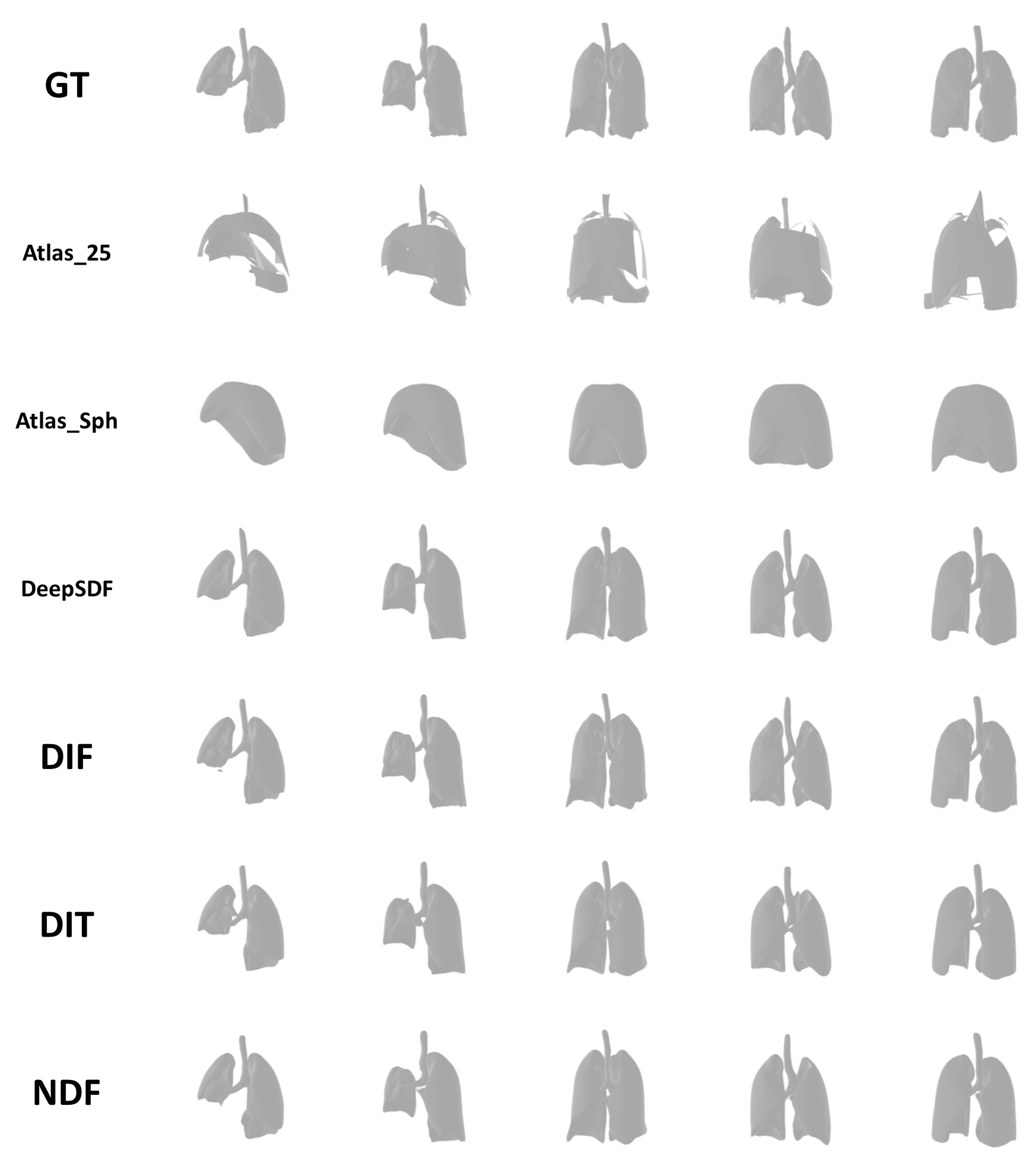}
        \caption{Lung}
        \label{fig:lung_train_rec}
    \end{subfigure}
    \caption{Seen Shape Representation Examples}
    \label{fig:train_rec}
\end{figure*}

\begin{figure*}
\centering
    \begin{subfigure}[b]{\columnwidth}
        \centering
        \includegraphics[width=\columnwidth]{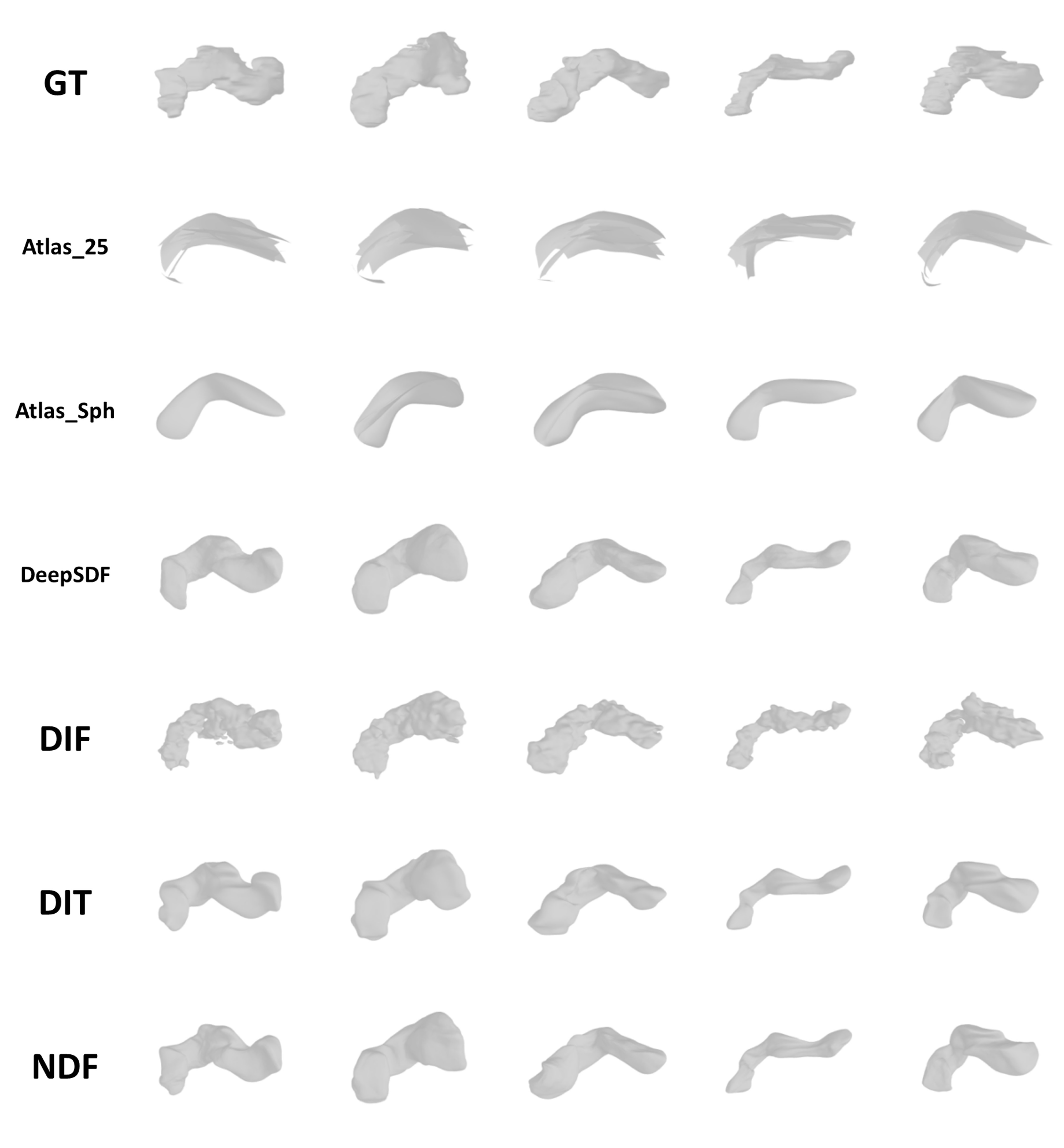}
        \caption{Pancreas}
        \label{fig:pancreas_test_rec}
    \end{subfigure}
    \hfill
    \begin{subfigure}[b]{\columnwidth}
        \centering
        \includegraphics[width=\columnwidth]{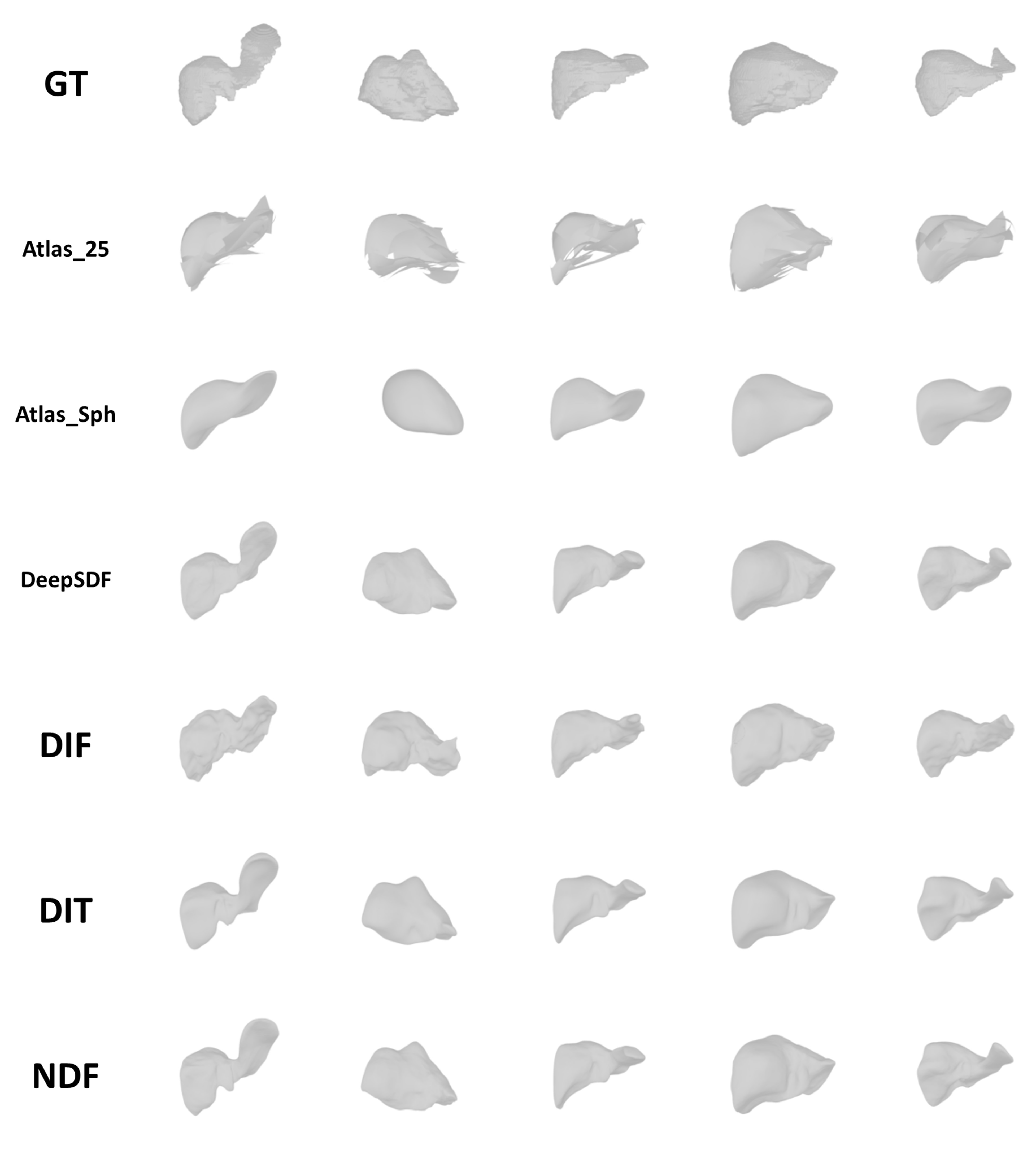}
        \caption{Liver}
        \label{fig:liver_test_rec}
    \end{subfigure}
    \vfill
    \begin{subfigure}[b]{\columnwidth}
        \centering
        \includegraphics[width=\columnwidth]{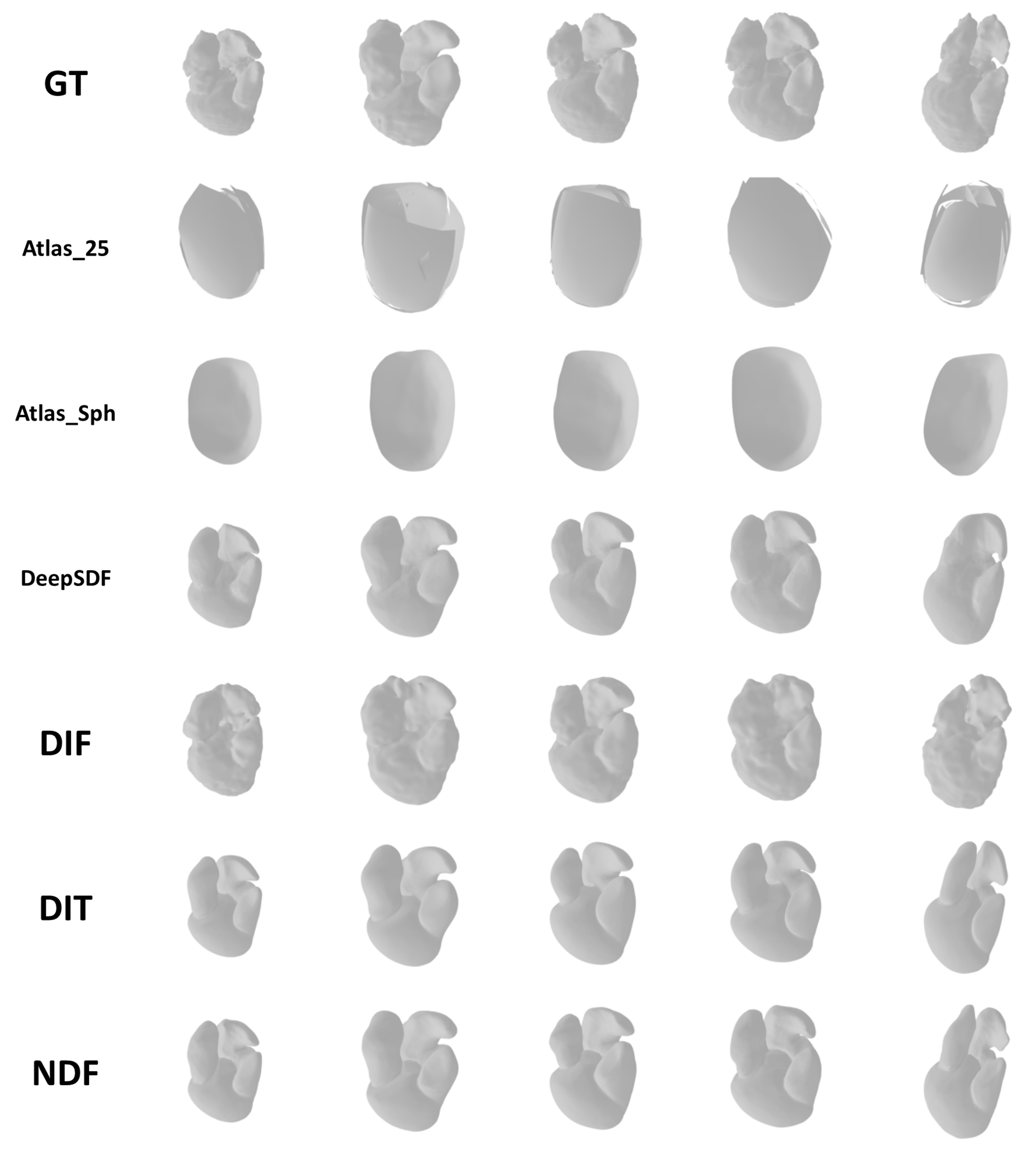}
        \caption{Heart}
        \label{fig:mmwhs_test_rec}
    \end{subfigure}
    \hfill
    \begin{subfigure}[b]{\columnwidth}
        \centering
        \includegraphics[width=\columnwidth]{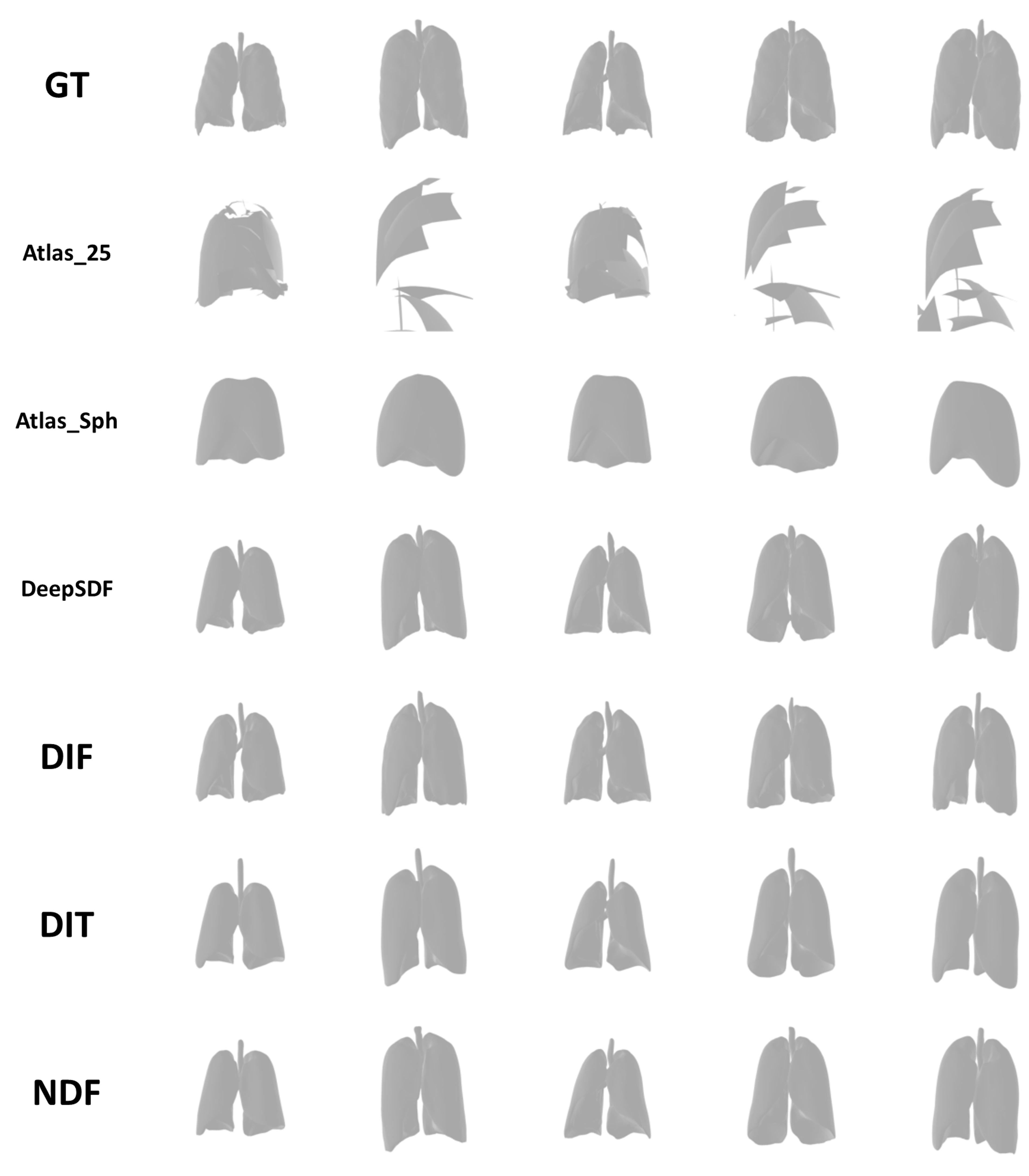}
        \caption{Lung}
        \label{fig:lung_test_rec}
    \end{subfigure}
    \caption{Unseen Shape Reconstruction Examples}
    \label{fig:test_rec}
\end{figure*}

In Fig.~\ref{fig:train_rec}, we select five cases for each class to demonstrate the seen shape representation results of six methods we compared in the main paper. In Fig.~\ref{fig:test_rec}, we select five cases for each class to demonstrate the unseen shape reconstruction results of six methods we compared in the main paper.

\paragraph{Seen and Unseen Shape Registration}
\begin{figure*}
\centering
    \begin{subfigure}[b]{\columnwidth}
        \centering
        \includegraphics[width=\columnwidth]{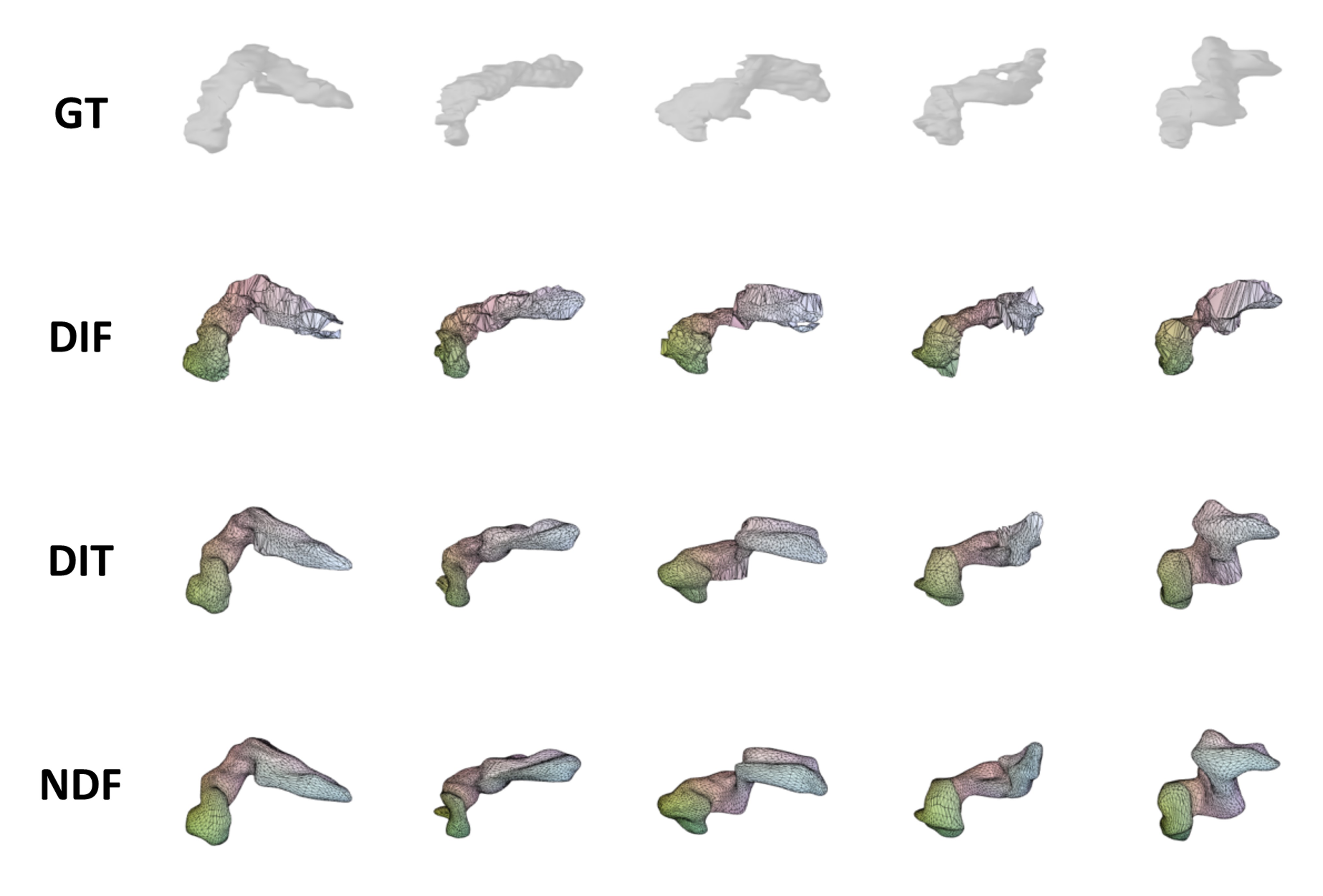}
        \caption{Pancreas}
        \label{fig:pancreas_train_2500_reg}
    \end{subfigure}
    \hfill
    \begin{subfigure}[b]{\columnwidth}
        \centering
        \includegraphics[width=\columnwidth]{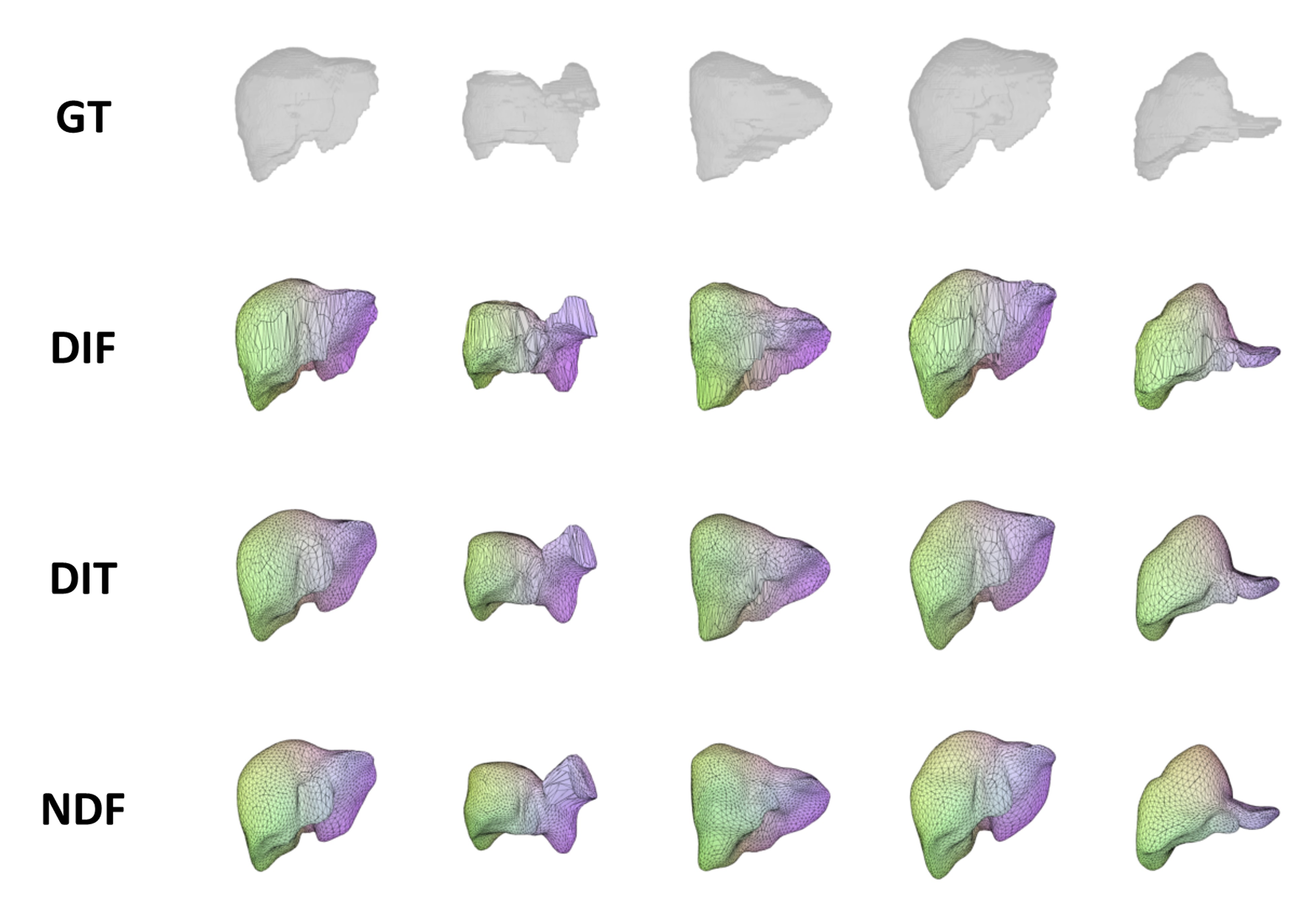}
        \caption{Liver}
        \label{fig:liver_train_2500_reg}
    \end{subfigure}
    \vfill
    \begin{subfigure}[b]{\columnwidth}
        \centering
        \includegraphics[width=\columnwidth]{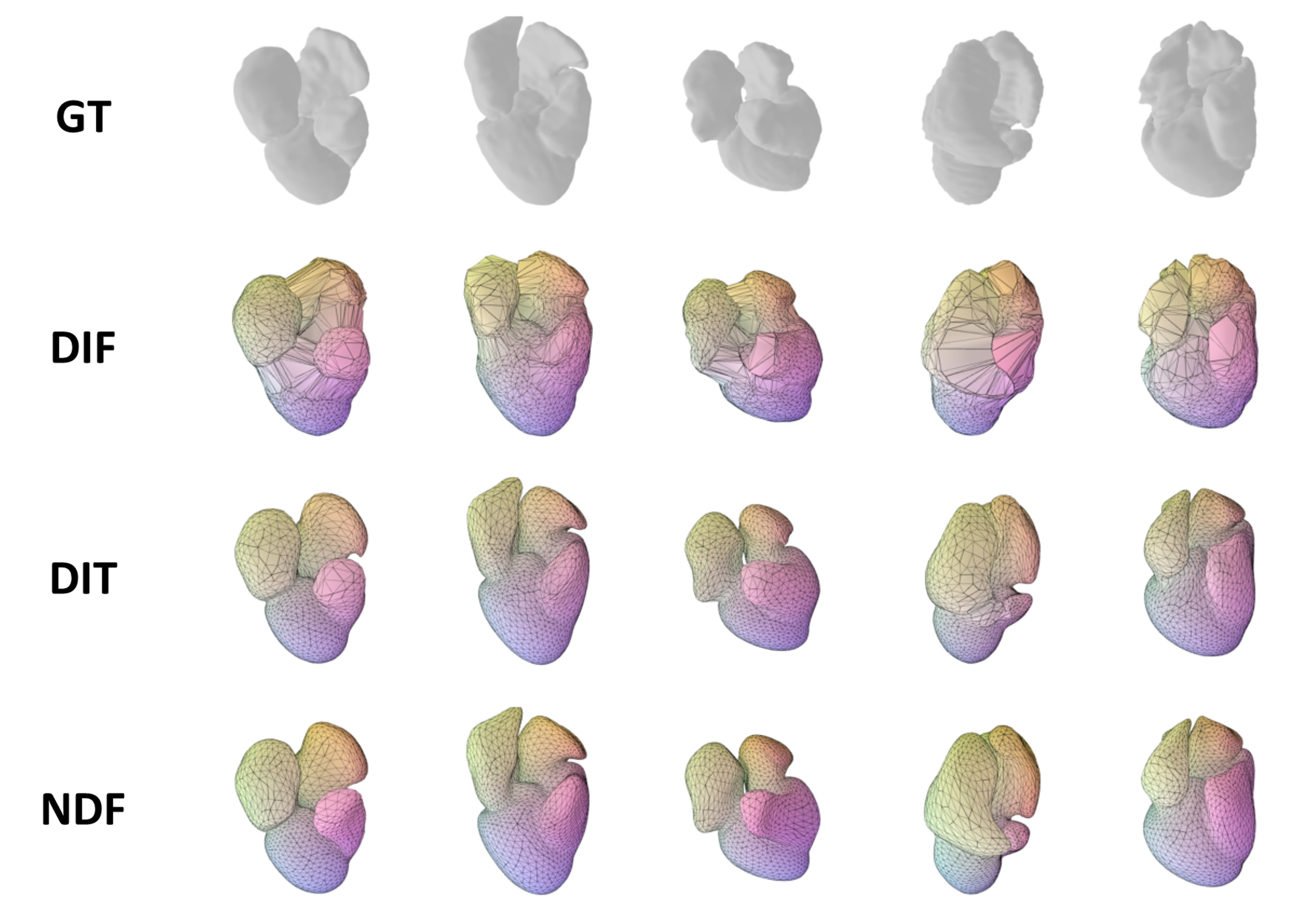}
        \caption{Heart}
        \label{fig:mmwhs_train_2500_reg}
    \end{subfigure}
    \hfill
    \begin{subfigure}[b]{\columnwidth}
        \centering
        \includegraphics[width=\columnwidth]{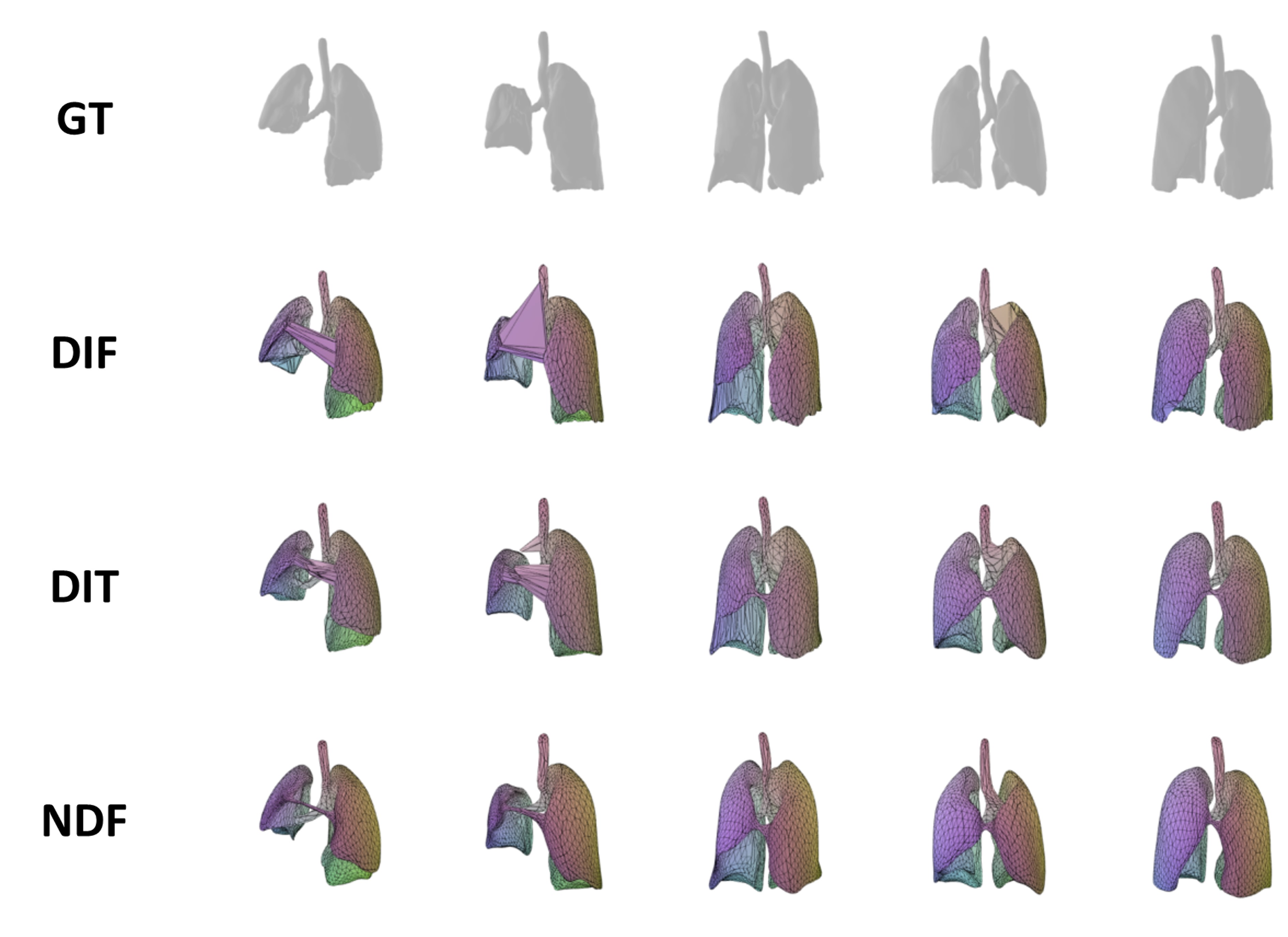}
        \caption{Lung}
        \label{fig:lung_train_2500_reg}
    \end{subfigure}
    \caption{Seen Shape Registration Examples with 2500-vertex Template Shapes}
    \label{fig:train_2500_reg}
\end{figure*}

\begin{figure*}
\centering
    \begin{subfigure}[b]{\columnwidth}
        \centering
        \includegraphics[width=\columnwidth]{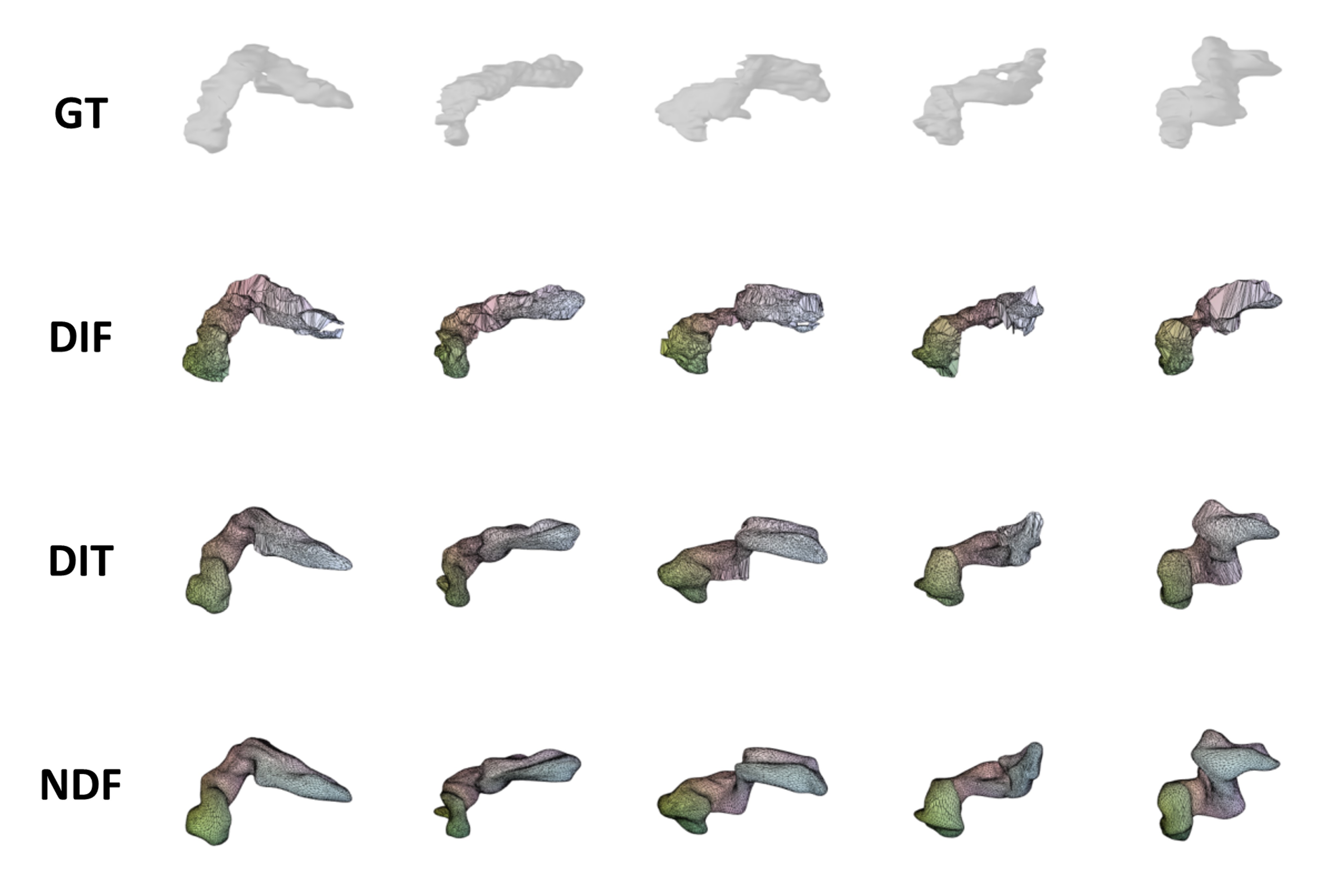}
        \caption{Pancreas}
        \label{fig:pancreas_train_5000_reg}
    \end{subfigure}
    \hfill
    \begin{subfigure}[b]{\columnwidth}
        \centering
        \includegraphics[width=\columnwidth]{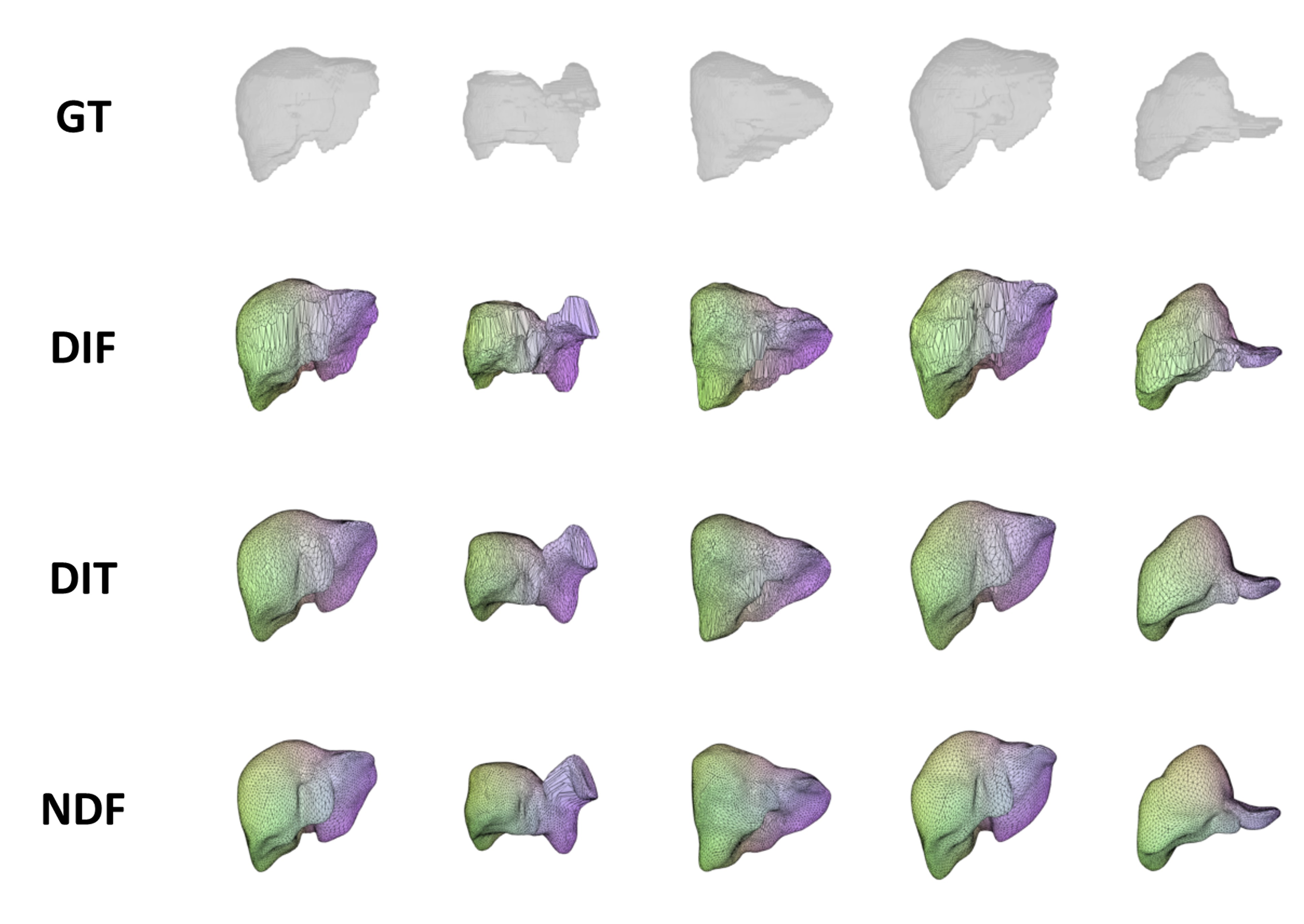}
        \caption{Liver}
        \label{fig:liver_train_5000_reg}
    \end{subfigure}
    \vfill
    \begin{subfigure}[b]{\columnwidth}
        \centering
        \includegraphics[width=\columnwidth]{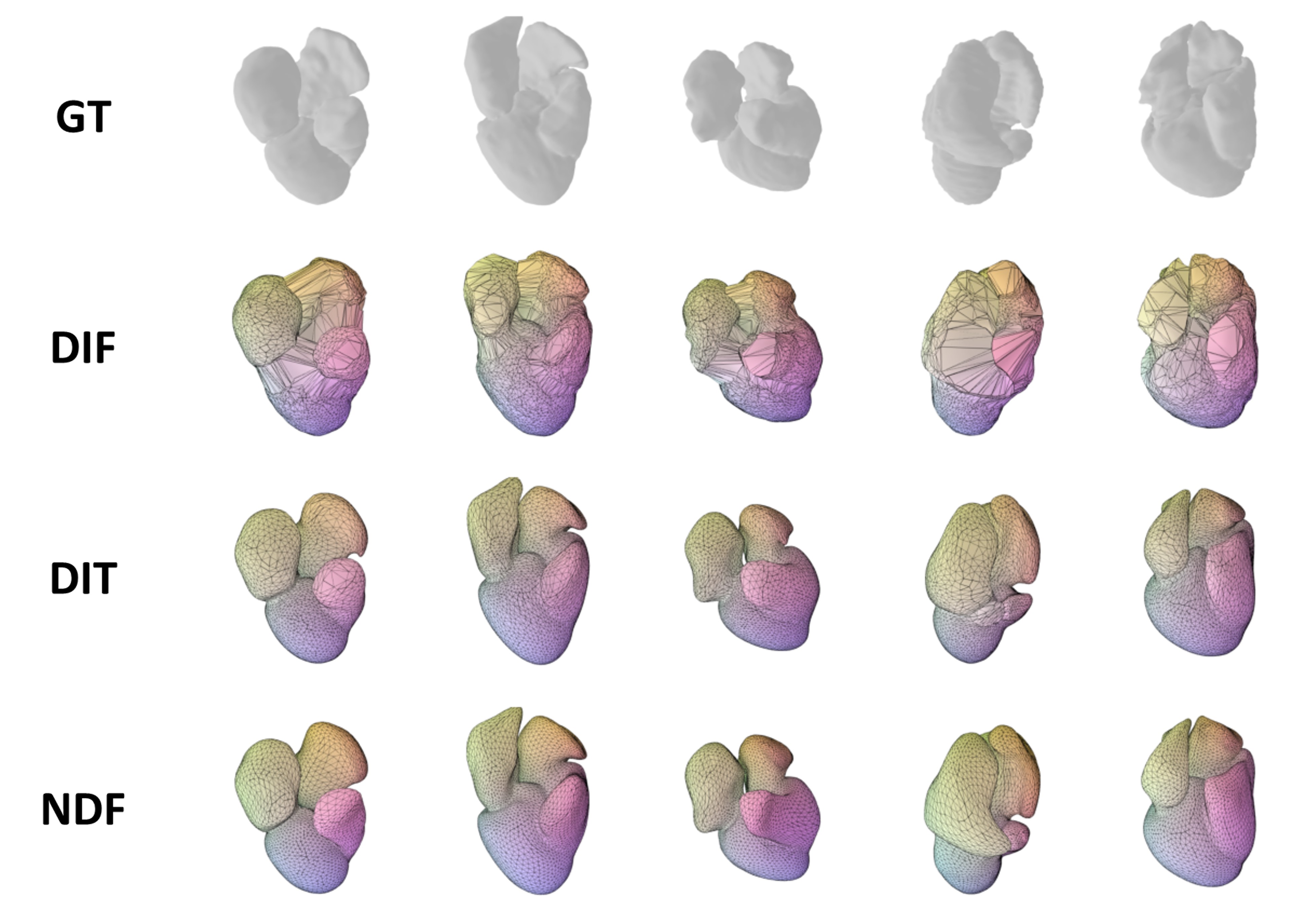}
        \caption{Heart}
        \label{fig:mmwhs_train_5000_reg}
    \end{subfigure}
    \hfill
    \begin{subfigure}[b]{\columnwidth}
        \centering
        \includegraphics[width=\columnwidth]{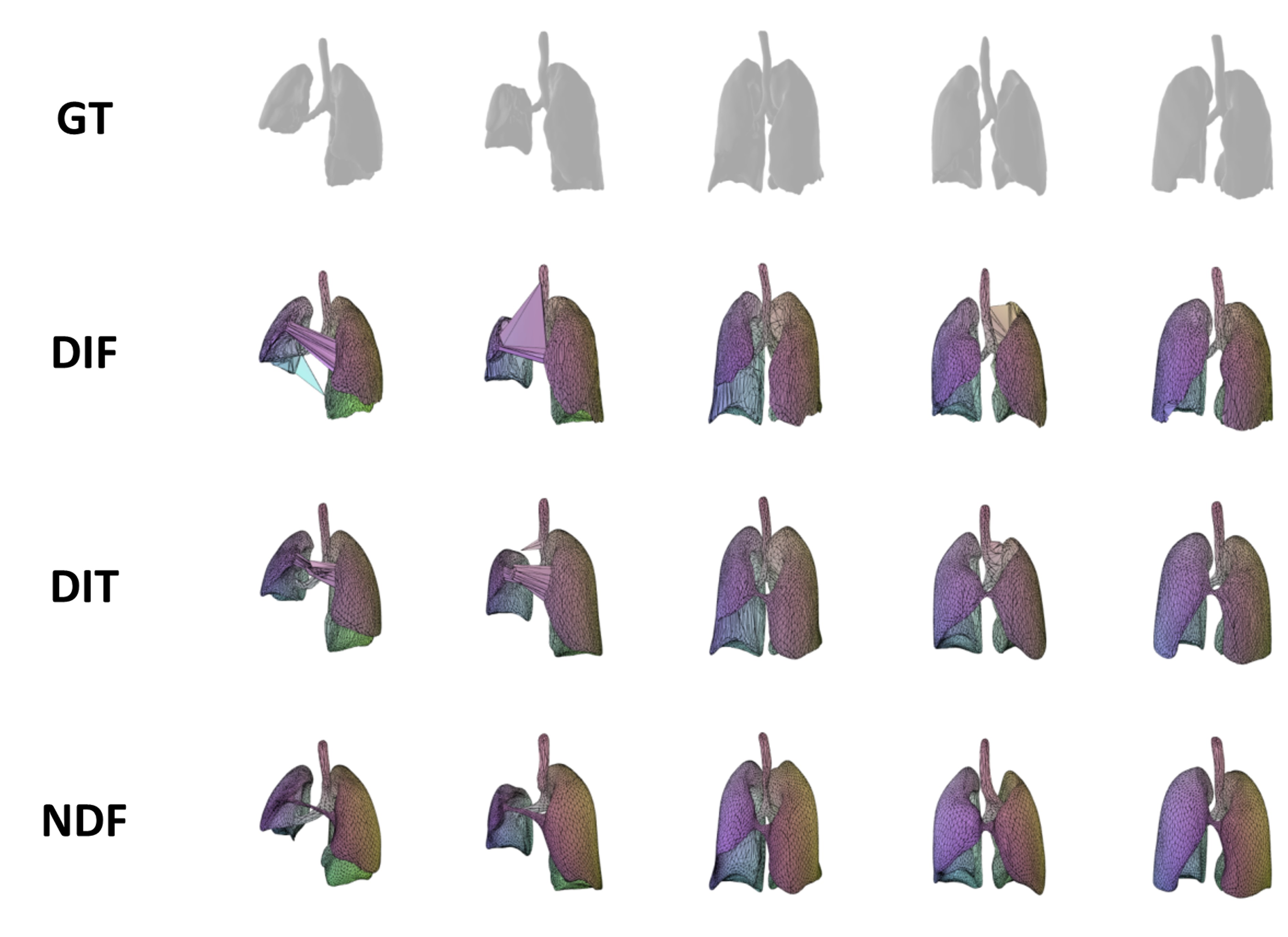}
        \caption{Lung}
        \label{fig:lung_train_5000_reg}
    \end{subfigure}
    \caption{Seen Shape Registration Examples with 5000-vertex Template Shapes}
    \label{fig:train_5000_reg}
\end{figure*}

\begin{figure*}
\centering
    \begin{subfigure}[b]{\columnwidth}
        \centering
        \includegraphics[width=\columnwidth]{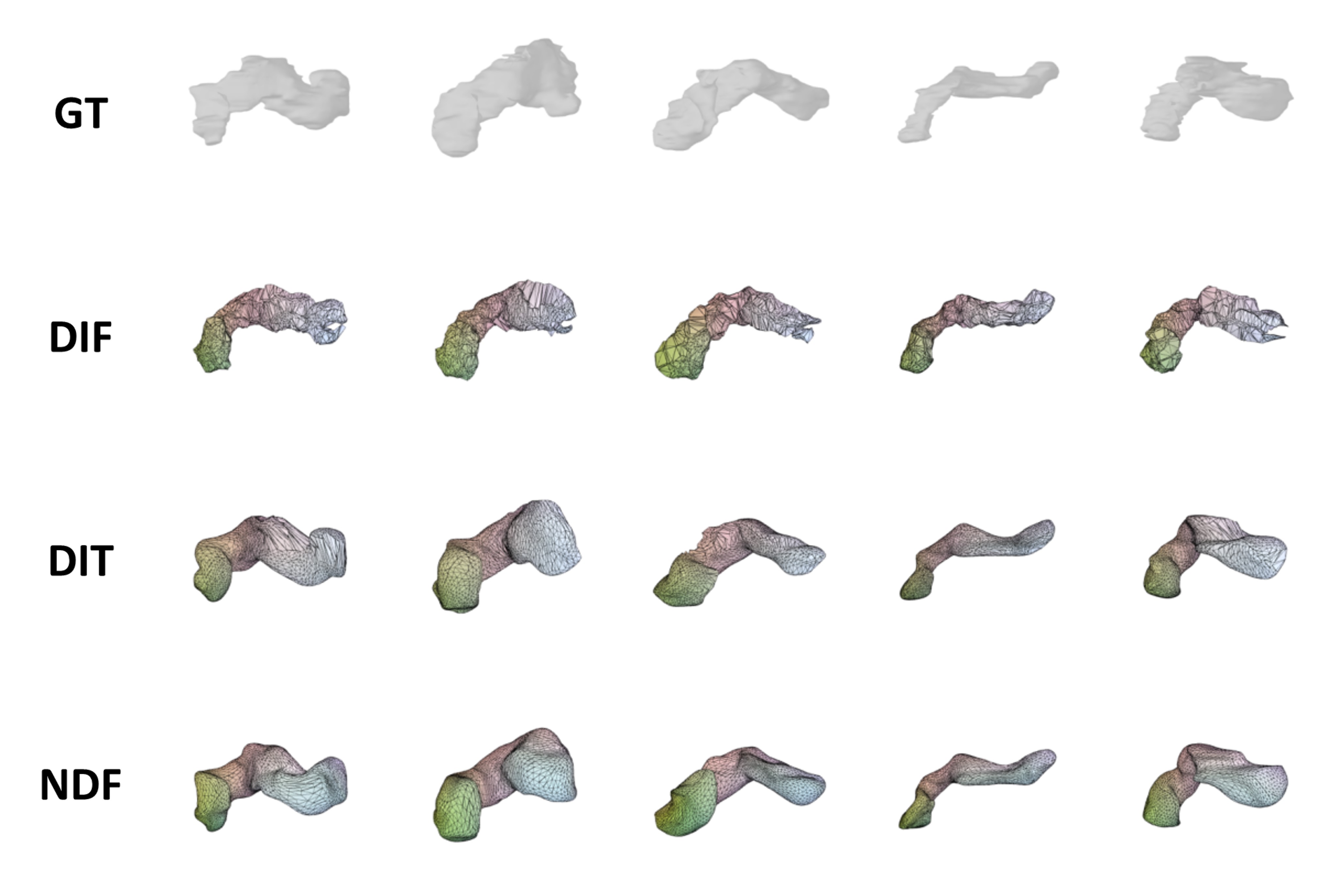}
        \caption{Pancreas}
        \label{fig:pancreas_test_2500_reg}
    \end{subfigure}
    \hfill
    \begin{subfigure}[b]{\columnwidth}
        \centering
        \includegraphics[width=\columnwidth]{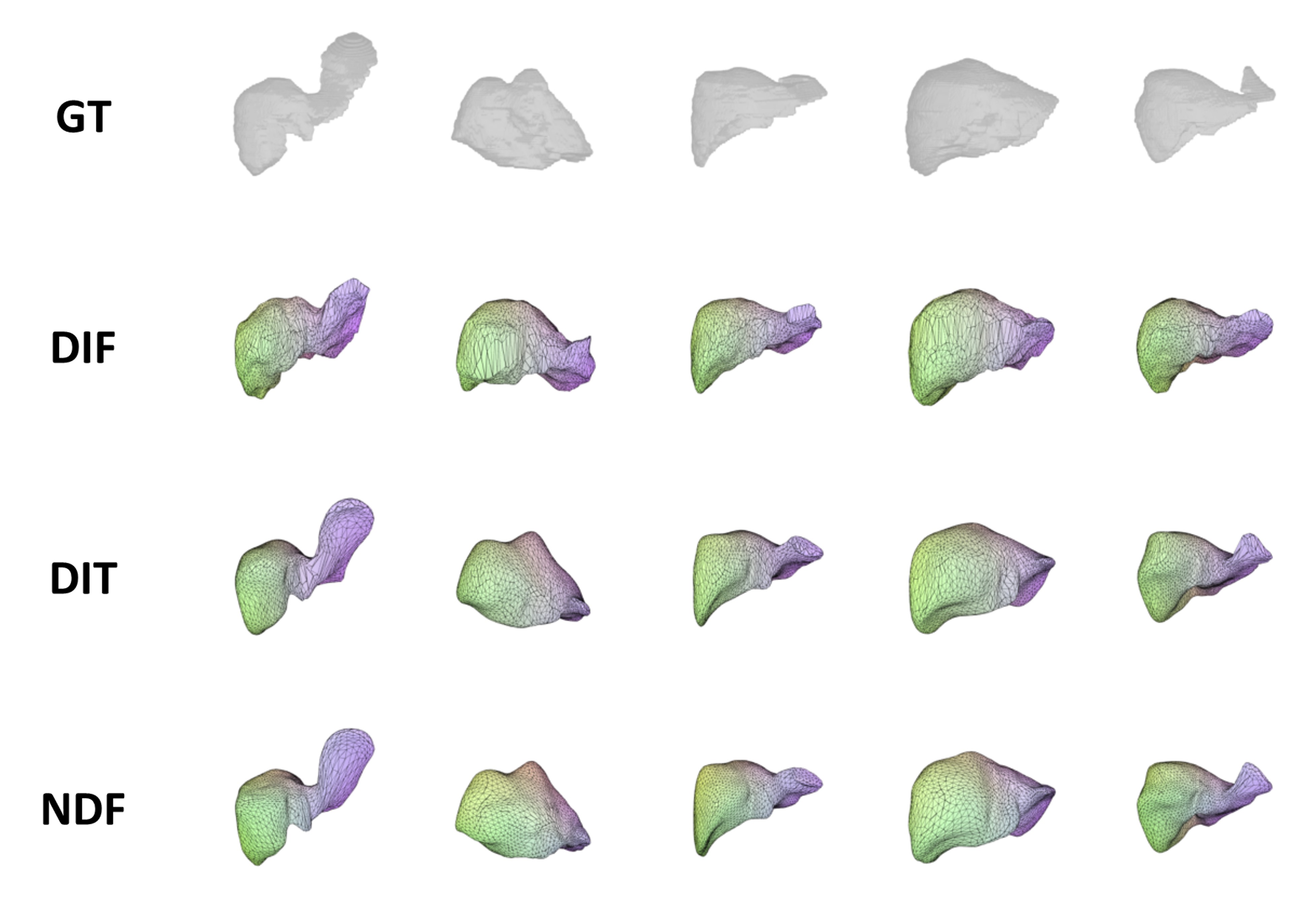}
        \caption{Liver}
        \label{fig:liver_test_2500_reg}
    \end{subfigure}
    \vfill
    \begin{subfigure}[b]{\columnwidth}
        \centering
        \includegraphics[width=\columnwidth]{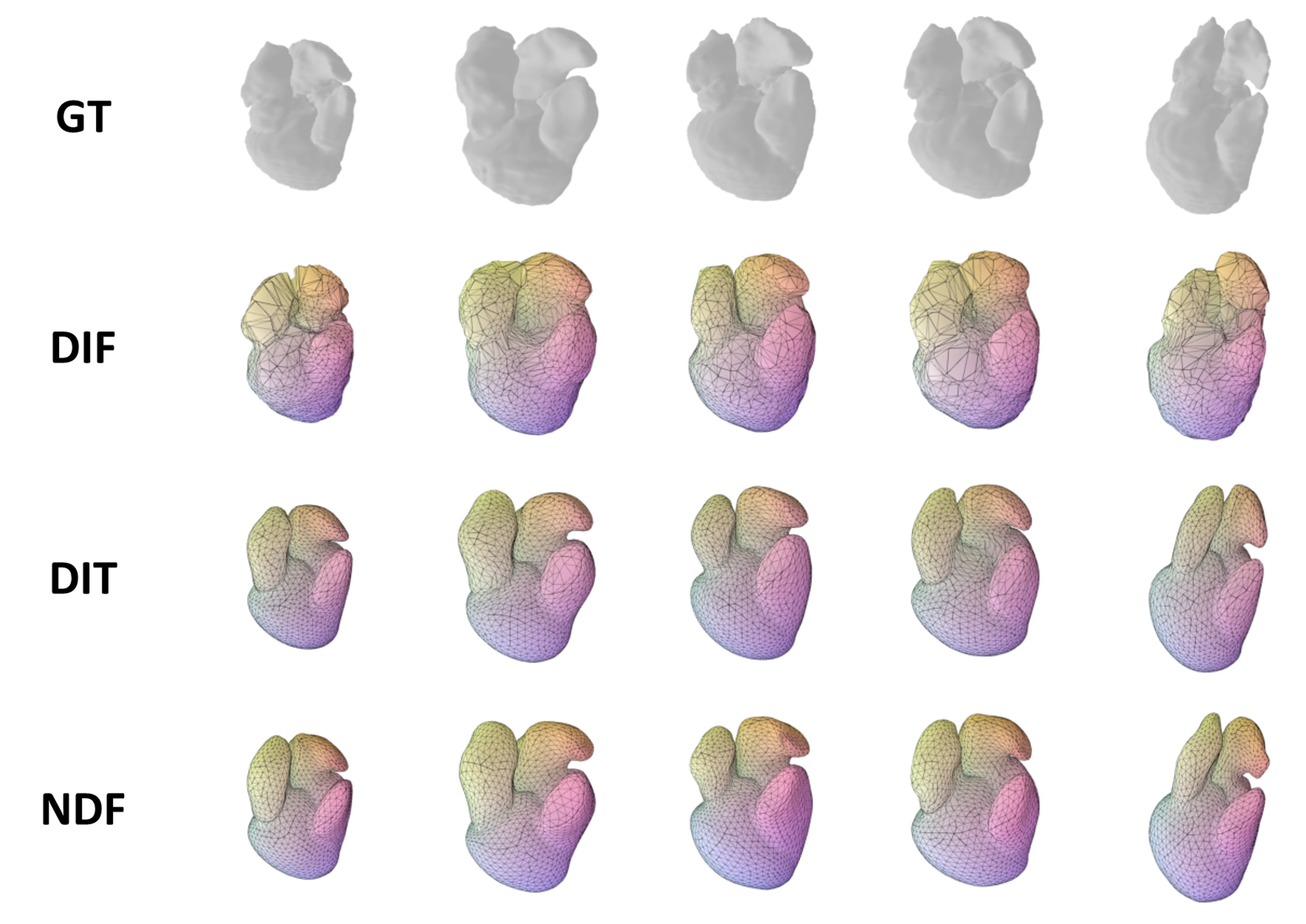}
        \caption{Heart}
        \label{fig:mmwhs_test_2500_reg}
    \end{subfigure}
    \hfill
    \begin{subfigure}[b]{\columnwidth}
        \centering
        \includegraphics[width=\columnwidth]{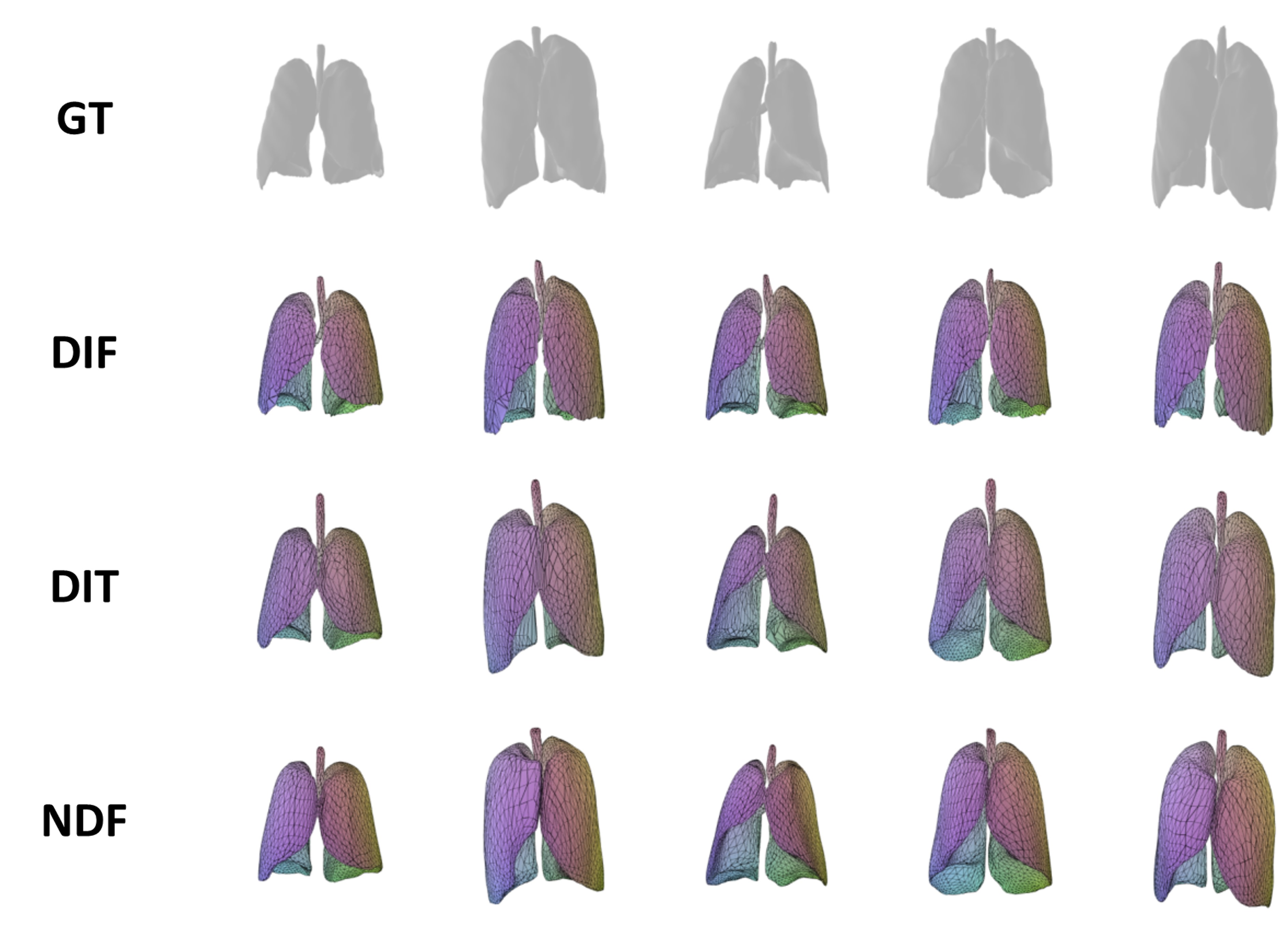}
        \caption{Lung}
        \label{fig:lung_test_2500_reg}
    \end{subfigure}
    \caption{Unseen Shape Registration Examples with 2500-vertex Template Shapes}
    \label{fig:test_2500_reg}
\end{figure*}

\begin{figure*}
\centering
    \begin{subfigure}[b]{\columnwidth}
        \centering
        \includegraphics[width=\columnwidth]{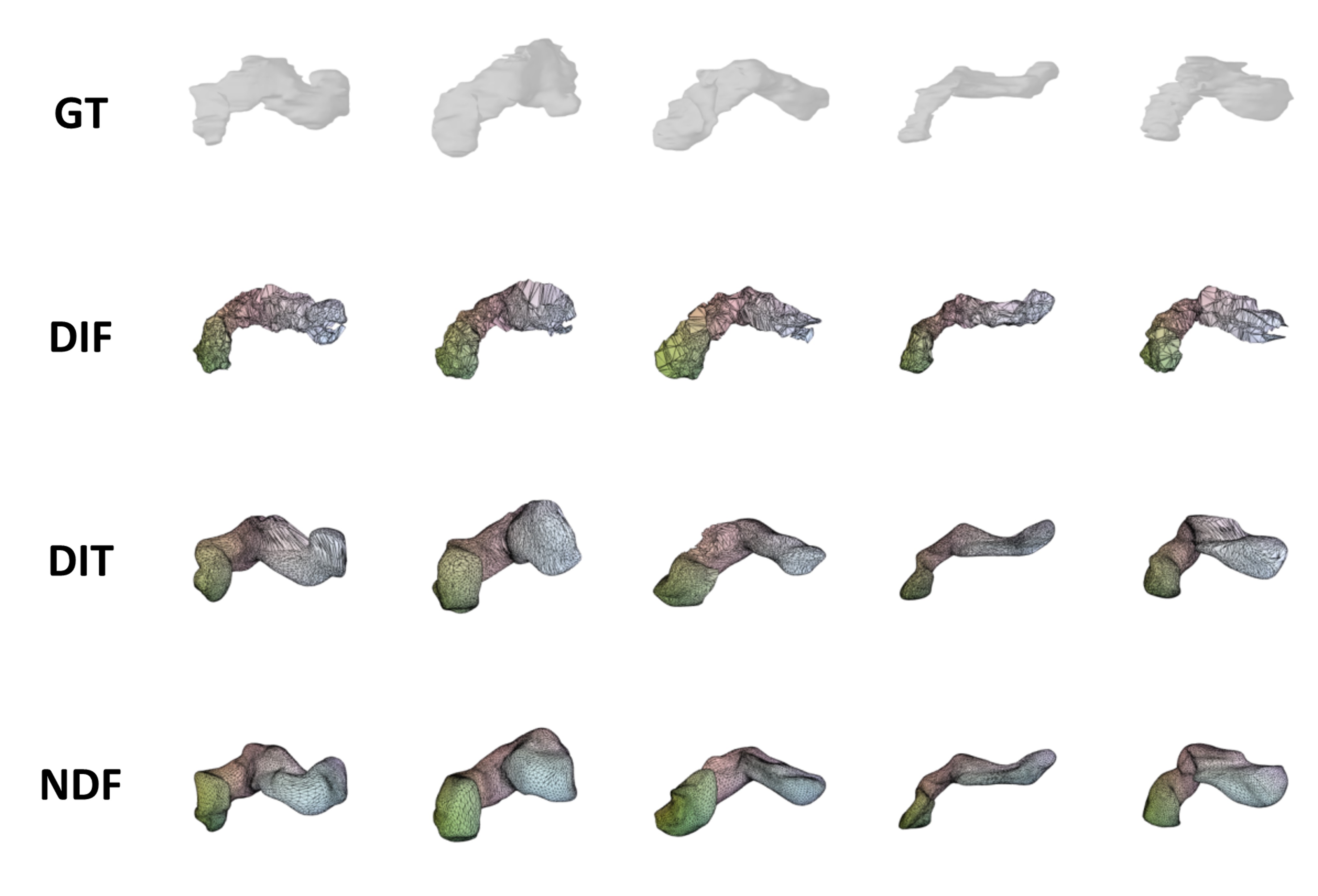}
        \caption{Pancreas}
        \label{fig:pancreas_test_5000_reg}
    \end{subfigure}
    \hfill
    \begin{subfigure}[b]{\columnwidth}
        \centering
        \includegraphics[width=\columnwidth]{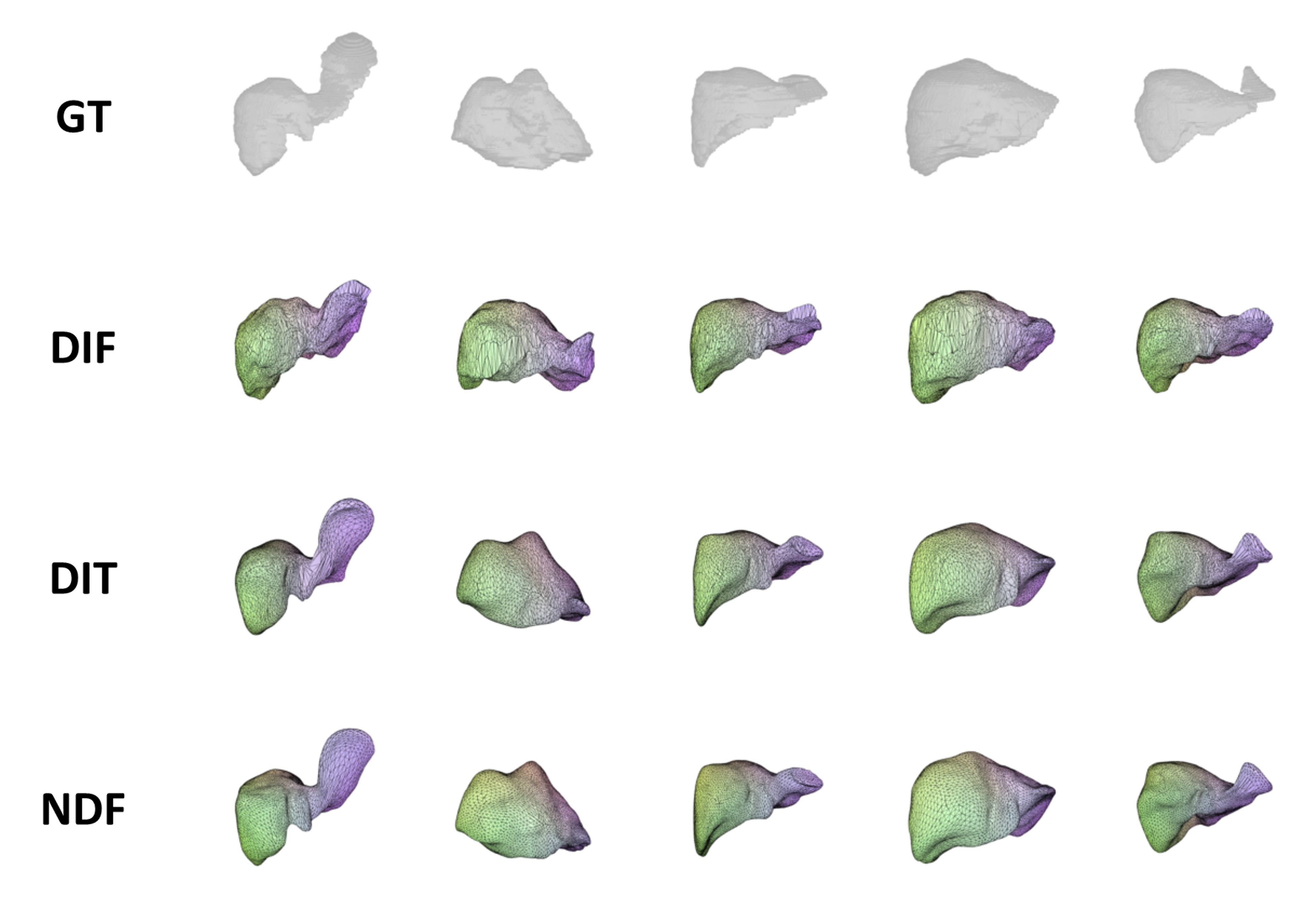}
        \caption{Liver}
        \label{fig:liver_test_5000_reg}
    \end{subfigure}
    \vfill
    \begin{subfigure}[b]{\columnwidth}
        \centering
        \includegraphics[width=\columnwidth]{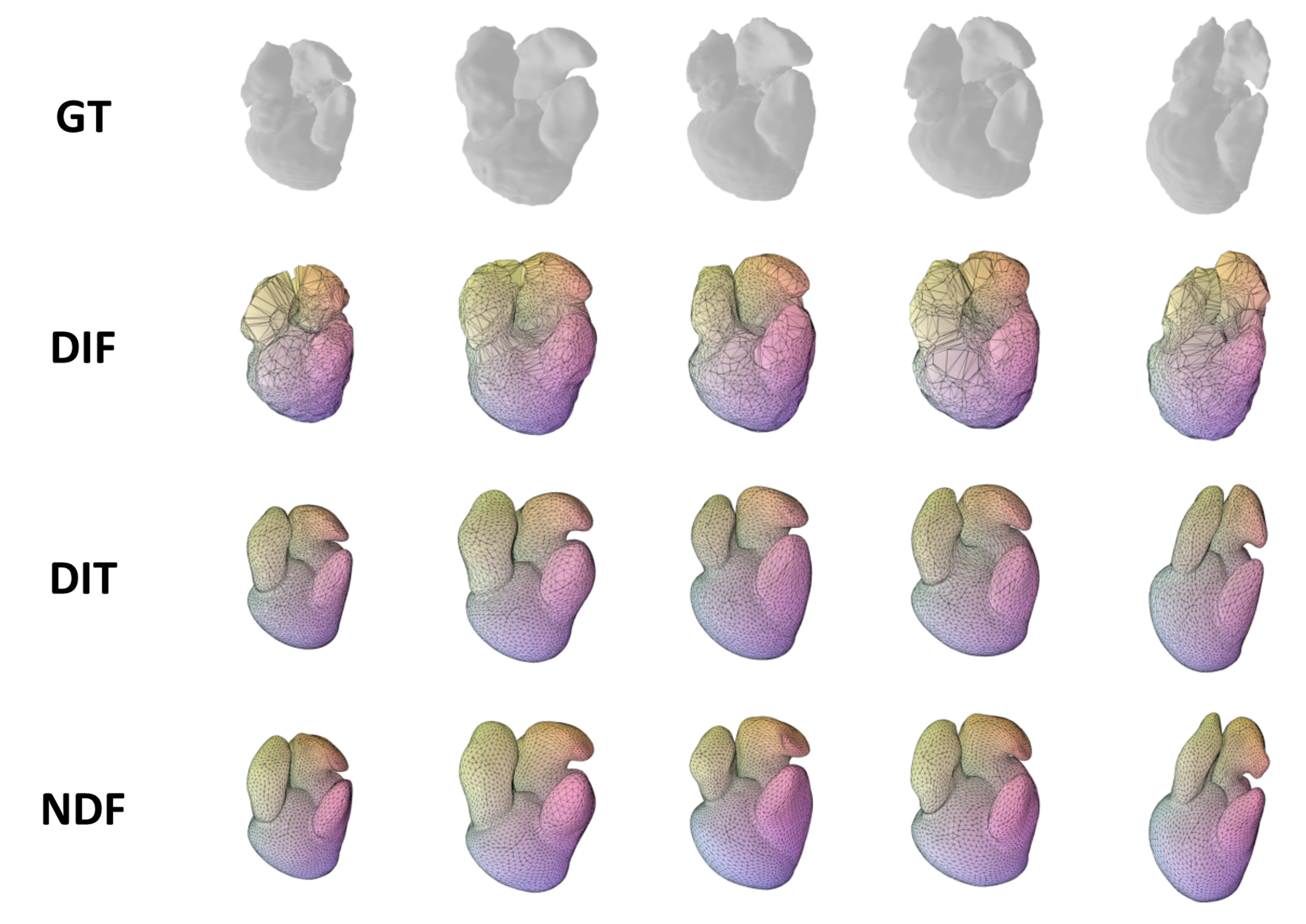}
        \caption{Heart}
        \label{fig:mmwhs_test_5000_reg}
    \end{subfigure}
    \hfill
    \begin{subfigure}[b]{\columnwidth}
        \centering
        \includegraphics[width=\columnwidth]{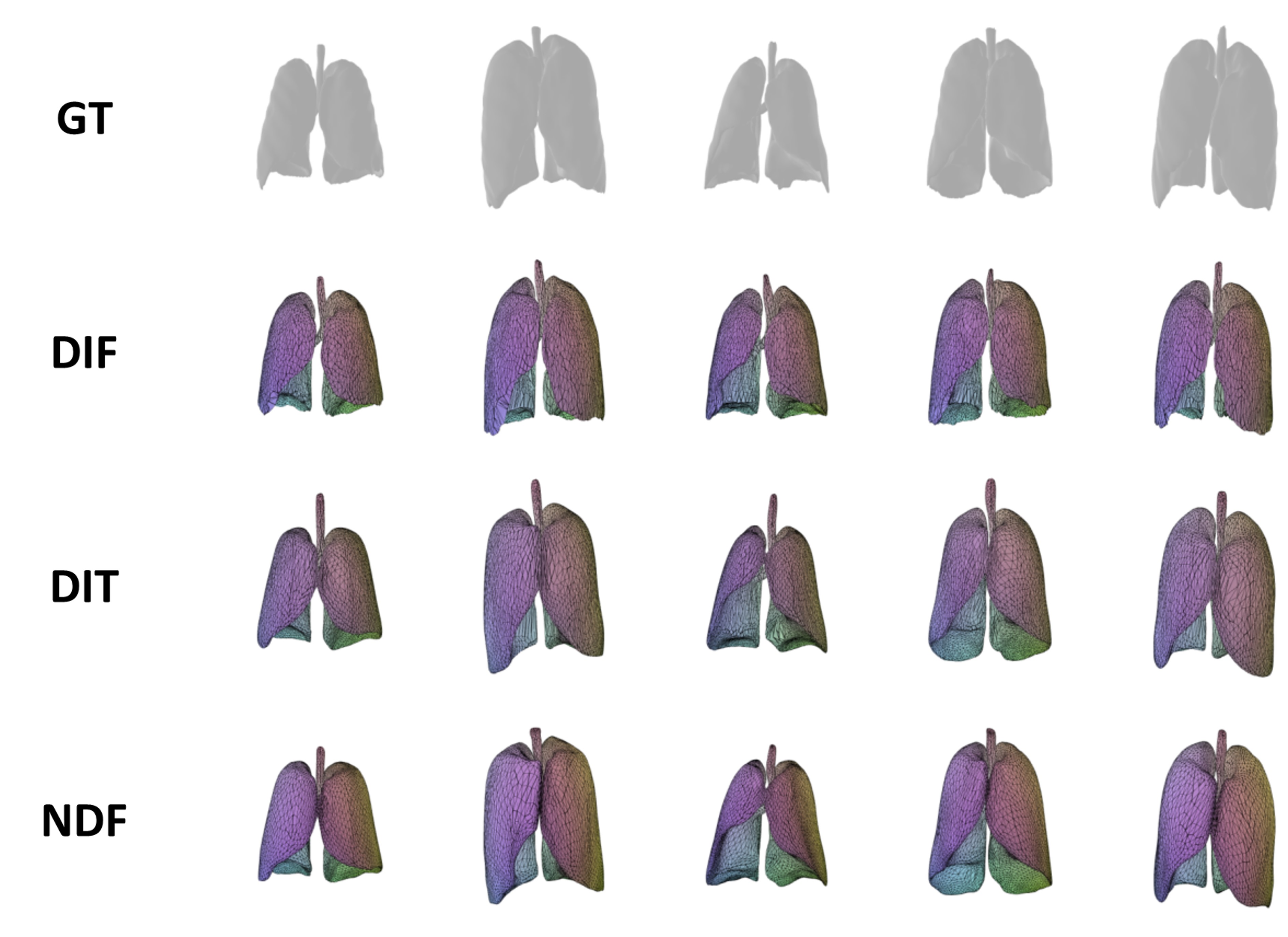}
        \caption{Lung}
        \label{fig:lung_test_5000_reg}
    \end{subfigure}
    \caption{Unseen Shape Registration Examples with 5000-vertex Template Shapes}
    \label{fig:test_5000_reg}
\end{figure*}

We sample 5 training cases and 5 test cases for the supplementary visual results. From Fig.~\ref{fig:train_2500_reg}-\ref{fig:test_5000_reg}, we can see our NDF is consistently better than the other two methods regarding both accuracy and fidelity when doing shape registration with verying-topology template shapes.



\end{document}